\documentclass{article}
\usepackage[pdftex]{graphicx}
\usepackage{amsmath}
\usepackage{placeins}
\usepackage{wrapfig}
\usepackage{caption}
\PassOptionsToPackage{numbers, compress}{natbib}

 \usepackage[main, final]{neurips_2026}

\usepackage[utf8]{inputenc} 
\usepackage[T1]{fontenc}    
\usepackage[hidelinks]{hyperref}
\usepackage{url}            
\usepackage{booktabs}       
\usepackage{amsfonts}       
\usepackage{nicefrac}       
\usepackage{microtype}      
\usepackage{xcolor}         

\title{\fontsize{15}{18}\selectfont{Do Pathology Foundation Models Encode Disease Progression? \\ A Pseudotime Analysis of Visual Representations}}

\author{
  Pritika Vig\\
  Massachusetts Institute of Technology, Cambridge, MA, USA \\
  Department of Data Science, Dana-Farber Cancer Institute, Boston, MA, USA \\
  \texttt{pritikav@mit.edu} \\
  \And
  Ren-Chin Wu \\
  Department of Pathology,
  Dana-Farber Cancer Institute, Boston, MA, USA \\
  \texttt{renchin\_wu@dfci.harvard.edu} \\
  \And
  William Lotter \\
  Department of Data Science, Dana-Farber Cancer Institute, Boston, MA, USA \\
  Brigham and Women's Hospital, Boston, MA, USA \\
  Harvard Medical School, Boston, MA, USA \\
  \texttt{lotterb@ds.dfci.harvard.edu}
}

\begin{document}

\maketitle

\begingroup
\renewcommand{\thefootnote}{}
\footnotetext{Code available at \url{https://github.com/lotterlab/cpath_dpt_progression}}
\endgroup

\begin{abstract}
  Vision foundation models trained on discretely sampled images achieve strong performance on classification benchmarks, yet whether their representations encode the continuous processes underlying their training data remains unclear. This question is especially pertinent in computational pathology, where we posit that models whose latent representations implicitly capture ordinal disease progression may better reflect underlying biology, support more robust generalization, and enable quantitative analyses of features associated with disease transitions. Using diffusion pseudotime, a method developed to infer developmental trajectories from single-cell transcriptomics, we probe whether foundation models organize disease states along coherent progression directions in representation space. Across four cancer progressions and six models, we find that all pathology-specific models recover trajectory orderings significantly exceeding null baselines, with pathology vision-only models achieving the highest average fidelities. Model rankings by trajectory fidelity on reference diseases predict few-shot classification performance on held-out diseases ($\rho = 0.91$), and exploratory analysis shows cell-type composition varies along inferred trajectories in patterns consistent with known stromal remodeling. Together, these results suggest that vision foundation models can implicitly learn to represent continuous processes from independent static observations, and that trajectory fidelity provides a complementary measure of representation quality beyond downstream performance. While demonstrated in pathology, this framework could be applied to other domains where continuous processes are observed through static snapshots.
\end{abstract}

\section{Introduction}
\label{sec:introduction}

Vision foundation models for computational pathology (CPath) now achieve state-of-the-art performance across diverse diagnostic tasks, from tumor classification to mutation prediction \citep{chen2024uni, zimmermann2024virchow, xu2024gigapath, lu2024conch, xiang2025musk}. These models process hematoxylin and eosin (H\&E) stained whole slide images (WSIs) by dividing them into smaller image patches and generating embeddings for each patch. The performance of these embeddings is primarily assessed using discrete classification benchmarks \citep{kang2023benchmarking, campanella2025clinical, bareja2025comprehensive}.

However, relying exclusively on downstream performance benchmarks provides insufficient understanding of both the ability to generalize in clinical settings and the extent to which foundation model representations encode underlying disease biology. Methods targeting the latter especially have the potential to advance our understanding of connections between histological features and patient outcomes. The importance of understanding foundation model representations is underscored by recent findings highlighting their fragility under distribution shifts and stain variations \citep{tizhoosh2025robustness, dejong2025distribution, komen2024batch, mishra2026}.

Robust pathology foundation models will require representations that capture true biological structure. One natural test of this quality is disease progression: while downstream benchmarks typically assess discrete classification, disease itself progresses continuously across ordinal states. Understanding what these transitions look like histologically is particularly important in oncology, as it can inform new treatment targets and risk stratification strategies. More broadly, the question of whether vision models implicitly learn to represent continuous processes from discretely sampled images extends to many other biomedical and natural image domains.  

Given its importance, methods have been developed to infer temporal ordering from independent biological samples in other applications. Among these, diffusion pseudotime (DPT; \citep{haghverdi2016diffusion}) reconstructs developmental trajectories from single-cell transcriptomics, where input features are gene expression measurements with direct biological meaning. DPT constructs a nearest-neighbor graph over samples and computes transition probabilities via random walks, yielding a pseudotime coordinate that orders cells along a continuous process.

Here, we introduce a pseudotime-based framework to study the opaque learned representations of CPath foundation models, with the intuition that if DPT recovers correct disease progression class ordering from these embeddings, it provides evidence that the representation implicitly encodes progression-relevant structure. Recovering ordinal progression structure is a strictly stronger geometric criterion than achieving class separability, as a representation can cleanly separate labeled classes while arranging them in an order inconsistent with biological progression. This distinction enables our analysis to extend beyond discrete labels and probe within-class biological variation. We apply our framework to six foundation models and four disease progressions, and subsequently assess associations with downstream performance generalization and biological features. 

Our contributions are as follows:
\begin{itemize}
    \item We demonstrate that pathology foundation models implicitly encode 
    disease progression as low-dimensional manifolds, largely recovering temporal class ordering from static snapshots without supervision.
    \item We find trajectory fidelity correlates with few-shot 
    generalization performance: models preserving progression structure on reference 
    cohorts achieve higher classification task accuracy on held-out diseases (mean 
    $\rho = 0.91$), offering a potential signal for generalization.
    \item We provide evidence that the learned representations encode biological processes at sub-class granularity: cell-type composition varies along the inferred disease manifold in patterns consistent with known stromal remodeling, providing opportunities to understand biological changes more finely than discrete progression class labels.
\end{itemize}

\section{Related Work}
\label{sec:related_work}

\paragraph{Representation Analysis in Vision Models.}

Feature learning in convolutional neural networks (CNNs) has been extensively characterized through visualization and diagnostic analyses \citep{zeiler2014visualizing}; in contrast, understanding how vision transformers (ViTs) organize learned representations remains comparatively underdeveloped. Since all state-of-the-art pathology models evaluated in this work are ViT-based, this emerging literature is directly relevant to our analysis. Unlike CNNs, which enforce hierarchical feature abstraction through localized receptive fields and pooling, \citet{raghu2022visiontransformerslikeconvolutional} show that Vision Transformers exhibit highly uniform representation similarity across layers, with substantial overlap between early and late representations. This uniformity arises from early global information mixing via self-attention and strong residual connections that propagate features throughout the network. Importantly, representational uniformity at the layer level does not imply semantic uniformity at the neuron level: while entire layers remain highly similar in representation space, the distribution of concepts encoded by individual neurons shifts with depth. \citet{dorszewski2025colorsclassesemergenceconcepts} demonstrate that ViTs exhibit a gradual progression in neuron-level concept complexity, from colors and textures in early layers to object- and class-level semantics in later layers.

\paragraph{Trajectory Modeling.}

Disease progression is a continuous dynamic process, but CPath datasets largely consist of static snapshots of disjoint tissue samples captured at single timepoints. This mirrors a fundamental challenge in single-cell trajectory inference which Diffusion Pseudotime \citep[DPT;][]{haghverdi2016diffusion} was developed to address. DPT defines a progression ordering between data samples through a Markov diffusion process with transition probabilities that are based on pairwise feature similarities. By aggregating transition probabilities over random walks of all lengths, DPT induces a multiscale diffusion distance that approximates geodesic structure on the data manifold. Empirically, these diffusion-based orderings yield stable progression orderings that are more robust to local sampling density variation than distance-based approaches (e.g., Euclidean distance in feature space) \citep{haghverdi2016diffusion}.

\paragraph{Pathology Foundation Models and Benchmarking.}

Numerous CPath foundation models have recently been developed, which are trained in a self-supervised fashion using large-scale datasets of H\&E-stained slides. These models commonly operate at a patch-level to generate embeddings from cropped H\&E patches and are typically evaluated using downstream task benchmarks using discrete clinical prediction problems, often via linear probing on frozen embeddings \citep{kang2023benchmarking, campanella2025clinical, bareja2025comprehensive}. While these benchmarks provide broad coverage across tasks and cohorts, they assess discriminative performance between labeled states rather than whether learned representations preserve the continuous biological processes linking those states.

\section{Methods}
\label{sec:methods}

\begin{figure*}[t]
    \centering
    \includegraphics[width=\textwidth]{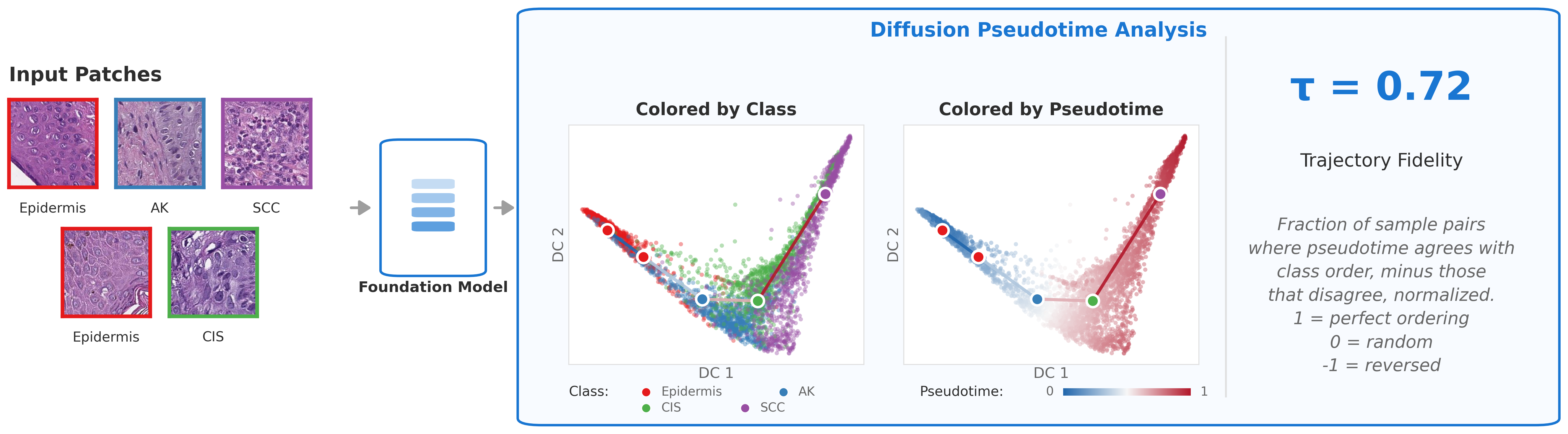}
    \caption{\textbf{Overview of trajectory fidelity evaluation.} H\&E patches representing disease states (here: skin squamous cell carcinoma progression from epidermis through actinic keratosis and carcinoma in situ to invasive SCC) are embedded by a foundation model (Virchow-2). We apply diffusion pseudotime analysis to the embedding space, visualized here via diffusion components (DC1, DC2). Left: patches colored by ground-truth class; Right: the same embeddings colored by inferred pseudotime. Trajectory fidelity ($\tau$, Kendall's rank correlation) quantifies agreement between pseudotime ordering and ground-truth ordinal class labels.}
    \label{fig:methods_overview}
\end{figure*}

\subsection{Diffusion Pseudotime}
\label{subsec:diffusion_pseudotime}

Given a set of $d$-dimensional feature vectors $\mathbf{X} = \{\mathbf{x}_1, \ldots, \mathbf{x}_n\}$ representing biological samples, diffusion pseudotime (DPT) identifies a relative ordering of the samples by computing transition probabilities between nearby points in feature space. Starting from a designated root sample $\mathbf{x}_{\text{root}}$, DPT aggregates these probabilities across random walks of all lengths, ultimately outputting a pseudotime coordinate $t_i = \text{dpt}(\mathbf{x}_{\text{root}}, \mathbf{x}_i) \in \mathbb{R}$ for each sample $i$ that quantifies its progression distance from the root. In the original single-cell genomics work, DPT successfully ordered 3,934 hematopoietic cells (42-dimensional gene expression features) from early precursors through intermediate states to mature blood cells, recovering the known developmental timeline~\citep{haghverdi2016diffusion}.

We adapt this approach to foundation model embeddings $\mathbf{X} = \{\mathbf{x}_1, \ldots, \mathbf{x}_n\}$ representing histopathology patches across disease states.
We build a transition matrix $\mathbf{T}$ by constructing a weighted $k$-nearest-neighbor graph over patch embeddings using cosine distance. To account for varying sampling densities on the manifold, we employ a locally scaled Gaussian kernel~\citep{haghverdi2016diffusion} as implemented in Scanpy~\citep{wolf2018scanpy}. The transition probability between patch features $\mathbf{x}_i$ and $\mathbf{x}_j$ is:

\begin{equation}
\begin{aligned}
T_{ij} &= \frac{K(\mathbf{x}_i, \mathbf{x}_j)}{Z_i}; \hspace{2pt}
K(\mathbf{x}_i, \mathbf{x}_j) = \sqrt{\frac{2\sigma_i\sigma_j}{\sigma_i^2+\sigma_j^2}}\,
\exp\left(-\frac{d(\mathbf{x}_i, \mathbf{x}_j)^2}{\sigma_i^2+\sigma_j^2}\right)
\end{aligned}
\end{equation}
where $d(\cdot,\cdot)$ is cosine distance, $\sigma_i$ is the local kernel width determined by the median distance from $\mathbf{x}_i$ to its $k$ nearest neighbors, and $Z_i = \sum_{j \in \mathcal{N}_k(i)} K(\mathbf{x}_i, \mathbf{x}_j)$ is the row-normalizing constant (applied after density normalization) ensuring $\sum_j T_{ij} = 1$.

For two patches indexed by $i$ and $j$, the DPT distance~\citep{haghverdi2016diffusion} is the Euclidean distance between their accumulated transition probabilities, $\text{dpt}(i,j) = \|\mathbf{M}(i,\cdot) - \mathbf{M}(j,\cdot)\|_2$, where $\mathbf{M}$ is the accumulated transition matrix summing over random walks of all lengths:
\begin{equation}
\mathbf{M} = \sum_{t=1}^{\infty} \tilde{\mathbf{T}}^t = (\mathbf{I} - \tilde{\mathbf{T}})^{-1} - \mathbf{I}
\end{equation}
Here $\tilde{\mathbf{T}} = \mathbf{T} - \psi_0\psi_0^T$ is the transition matrix with the stationary eigenspace (corresponding to eigenvalue $\lambda_0=1$) removed, with all spectral quantities computed as in Scanpy~\citep{wolf2018scanpy} from the symmetric form of $T$. In practice, $\mathbf{M}$ is approximated via spectral decomposition ~\citep{wolf2018scanpy}:
\begin{equation}
\mathbf{M} \approx \sum_{i=1}^{n_{\text{dc}}} \frac{\lambda_i}{1-\lambda_i} \psi_i \psi_i^T
\end{equation}
where $\{\lambda_i, \psi_i\}$ are the eigenvalue-eigenvector pairs of $\tilde{\mathbf{T}}$, ordered by decreasing eigenvalue magnitude and $n_{dc}$ is the number of diffusion components. Fixing a known root patch $r$, the pseudotime coordinate of patch $i$ is $t_i = \text{dpt}(r, i)$. In our experiments, we set $r$ as the medoid of embeddings from the earliest disease class. We compute DPT using the Scanpy~\citep{wolf2018scanpy} library. Hyperparameter selection and sensitivity analysis are detailed in Appendix~\ref{app:hyperparameters}.

\subsection{Trajectory Fidelity Metric}
\label{subsec:trajectory_fidelity}

To evaluate the agreement between the inferred diffusion pseudotime (continuous) and ground-truth disease classes (ordinal states), we computed Kendall’s Rank Correlation Coefficient ($\tau$)~\citep{haghverdi2016diffusion, kendall1938new}. This metric was used because it is the criterion under which DPT was originally validated~\citep{haghverdi2016diffusion}, it compares the order rather than the magnitude of pseudotime values, and it explicitly accounts for ties in the discrete class labels (via the $\tau_b$ variant) without assuming equal spacing between classes. A value of $\tau = 1$ indicates that the inferred pseudotime strictly preserves the biological order of classes in disease progression, whereas a value of 0 indicates random concordance. 

We note that class separability alone does not guarantee high $\tau$. Representations that cluster classes distinctly but arrange them inconsistently with progression direction would yield low $\tau$, whereas orderings that preserve the biological sequence along the embedding manifold would achieve $\tau \to 1$ (see Appendix~\ref{app:tau_vs_separability} for an illustrative simulation and an empirical example, as well as Section~\ref{subsubsec:permutation_test}).

\subsection{Disease Progression Cohorts}
\label{subsec:disease_cohorts}
We analyzed four biologically grounded cancer progressions using expert-annotated $224 \times 224$ pixel patches from the SPIDER dataset \citep{nechaev2025spidercomprehensivemultiorgansupervised}. SPIDER provides H\&E-stained patches at $20\times$ magnification from the HISTAI collection \citep{nechaev2025histai}, consisting of slides from multiple institutions and digitized using four different scanners, providing a robust dataset for our analysis. SPIDER was not included in training data for any foundation models we evaluate. Using the patch-level class labels, we defined four discrete progression trajectories based on established pathological models:

\textbf{Skin Squamous Cell Carcinoma (SCC).} \textit{Epidermis $\to$ Actinic keratosis $\to$ SCC in situ $\to$ Invasive SCC}.
This trajectory models classic UV-induced progression from precursor lesions to invasive malignancy \citep{Ratushny2012}.

\textbf{Colorectal Conventional Pathway.} \textit{Adenoma (low grade) $\to$ Adenoma (high grade) $\to$ Adenocarcinoma (low grade) $\to$ Adenocarcinoma (high grade)}. This follows the canonical adenoma-carcinoma sequence, representing gradual accumulation of genetic alterations from benign neoplasia to undifferentiated carcinoma \citep{Fearon1990}.

\textbf{Colorectal Serrated Pathway.} \textit{Hyperplastic polyp $\to$ Sessile serrated lesion (SSL) $\to$ Adenocarcinoma (high grade)}. This captures the alternative serrated pathway, distinct from the conventional sequence \citep{DePalma2019}. We note that SSL and high-grade adenocarcinoma span a relatively large morphological gap for which intermediate states can be assigned (e.g., SSL with dysplasia), but these more granular labels are not provided in the dataset. 

\textbf{Breast Ductal Carcinoma (BDC).} \textit{DCIS (low-grade) $\to$ DCIS (high-grade) $\to$ Invasive carcinoma}. While direct temporal evolution from low- to high-grade DCIS is one plausible trajectory, these grades can also arise via independent evolutionary pathways \citep{Hulahan2024}, and thus, at a minimum, this sequence reflects a progression of histological severity.

We standardized each disease progression to 1,000 patches per class using slide-aware sampling, which caps a single slide's contribution at 50 patches to mitigate intra-sample correlation and slide-level batch effects. This resulted in 4,000 patches for four-class cohorts (SCC, colorectal conventional) and 3,000 for colorectal serrated. Due to scarcity in the BDC dataset, the high-grade DCIS class was constrained to 714 patches from 39 available slides, yielding a total cohort of 2,714. Manifold robustness analysis confirmed that trajectory structure remains stable to these minor class density fluctuations (see Appendices~\ref{app:sampling} and~\ref{app:manifold_robustness}). Sensitivity analyses to slide selection and empirical vs. class-balanced sampling of patches additionally confirmed that our findings are robust to these hyperparameters (Appendix~\ref{app:resampling}).

\paragraph{Foundation Models.}
\label{subsec:foundation_models}

We evaluated six models spanning distinct pretraining paradigms, including one natural image baseline, DINOv2 \citep{oquab2024dinov2} 
(ViT-L/14 with registers, $d=1024$), and five CPath models. These comprise 
two methodological families: vision-only models trained with self-supervised 
learning, and vision-language models (VLMs) trained with vision-text contrastive learning. The 
vision-only CPath models—UNI-2 \citep{chen2024uni} (ViT-H/14, $d=1536$), 
Virchow-2 \citep{zimmermann2024virchow} (ViT-H/14, $d=1280$), and Prov-GigaPath 
\citep{xu2024gigapath} (ViT-g/14, $d=1536$)—all adapt DINOv2 self-distillation 
to histopathology. Broadly, these models differ in their scaling strategies and pathology-specific optimizations: UNI-2 focuses on architectural width (Custom ViT-H), Prov-GigaPath emphasizes massive real-world data scale (1.3B tiles from 171k slides), and Virchow-2 integrates multi-resolution training (5x-40x) with a domain-adapted DINOv2 objective. For vision-language models, CONCH \citep{lu2024conch} 
(ViT-B/16, $d=512$) uses a CoCa-style objective and MuSK \citep{xiang2025musk} 
(ViT-L/16, $d=1024$) uses a BEiT-3 \citep{wang2022imageforeignlanguagebeit} multimodal architecture. For all models, we extract [CLS] token embeddings from the patch encoder. For VLMs, we use embeddings from the vision encoder prior to text-alignment projection to isolate visual representations.

\paragraph{Statistical Analysis.}
Confidence intervals for $\tau$ were computed via stratified bootstrap resampling: patches were partitioned by ground-truth class, then resampled with replacement within each class while maintaining original class-wise sample sizes, followed by performing the DPT analysis on the bootstrapped sample. A null baseline was computed via randomly permuting class labels while maintaining the original DPT values. Bootstrap and permutation analyses used 100 iterations, which was picked to balance robustness with the high computational cost of recomputing the analysis. 

\section{Results}
\label{sec:results}

\begin{figure*}[t]
    \centering
    \includegraphics[width=\textwidth]{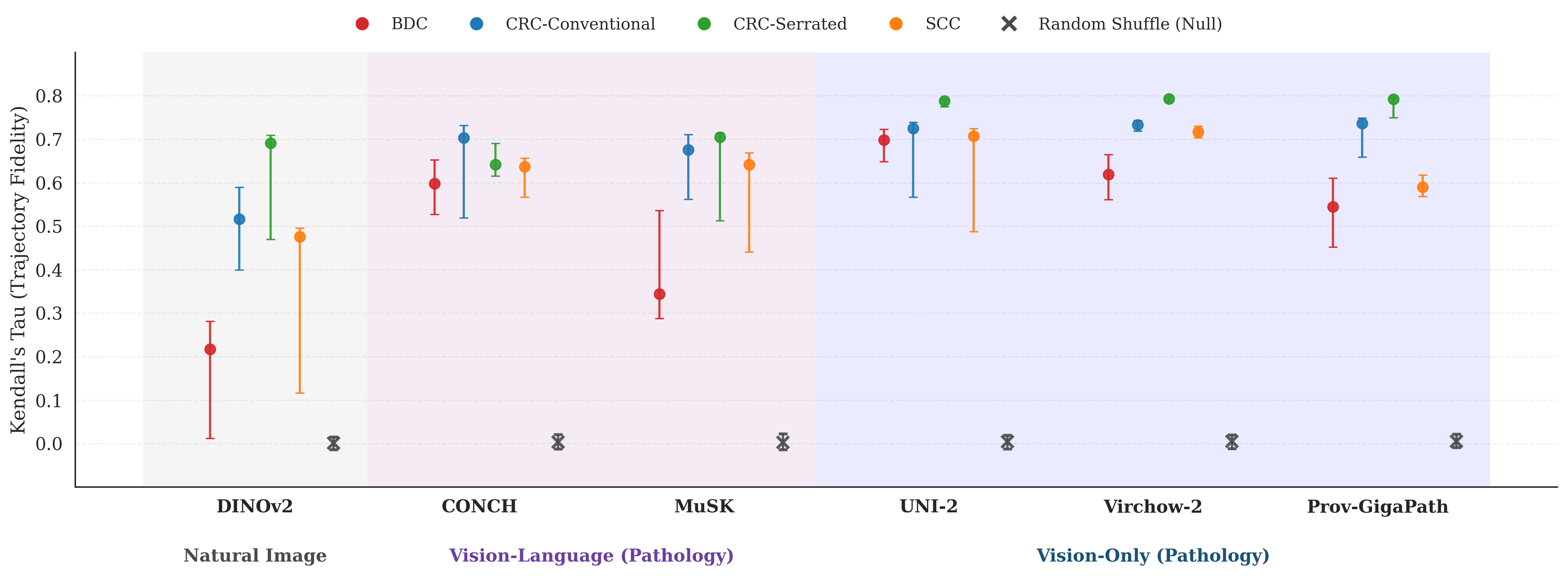}
    \caption{\textbf{Trajectory fidelity across foundation models and disease progressions.} Kendall's $\tau$ between diffusion pseudotime and ground-truth class labels for six foundation models across four cancer progressions. All CPath models significantly exceed the null baseline (gray ×, label-shuffled controls). UNI-2 and Virchow-2 show the highest average fidelity, while the natural image baseline (DINOv2) shows the lowest. Exact values reported in Appendix~\ref{app:traj_fidel_results}.}
    \label{fig:trajectory_fidelity}
    \vspace{-\intextsep}
\end{figure*}

\subsection{Foundation Model Embedding Manifolds Encode Disease Progression Trajectories}
\label{subsec:trajectory_results}

\subsubsection{Trajectory fidelity across models and diseases}
\label{subsubsec:trajectory_fidelity_results}

We evaluated trajectory fidelity across five computational pathology foundation models spanning two pretraining paradigms: vision-language (CONCH, MuSK), and vision-only (UNI-2, Virchow-2, Prov-GigaPath). We additionally compare to a natural image foundation model baseline (DINOv2). Figure~\ref{fig:trajectory_fidelity} reports Kendall's $\tau$ between inferred diffusion pseudotime and ground-truth disease state across all four progression cohorts.

All CPath models achieved trajectory fidelity significantly exceeding the label-shuffled null baseline. The separation from null confirms that patches with similar pseudotime values share disease states beyond what would be expected by chance, indicating that the embedding geometry places biologically adjacent states nearby. The natural image baseline (DINOv2) achieved moderate fidelity on colorectal progressions but failed to recover BDC progression ($\tau = 0.22$), approaching null performance. This suggests that general visual features capture some progression-relevant structure, but domain-specific training is important for consistent recovery across tissue types.

Among CPath models, UNI-2 and Virchow-2 showed the highest average fidelity across progressions (mean $\tau = 0.73$ and $0.72$, respectively), with their highest values on CRC-Serrated ($\tau = 0.79$ for both models). MuSK showed the largest spread in fidelity across progressions ($\tau = 0.34$ to $0.70$, a range of $0.36$ vs.\ $\leq 0.25$ for the other CPath models) and the lowest average fidelity (mean $\tau = 0.59$). The vision-only CPath models had higher average fidelity than either vision-language model, and a vision-only CPath model was the highest-performing model for each progression. However, we note that confidence intervals often overlapped between models on individual progressions, with the strongest pattern being the lower average fidelity of DINOv2 compared to all CPath models. 

The results presented in Figure~\ref{fig:trajectory_fidelity} are based on the class-balanced sampling of SPIDER patches as discussed in Section~\ref{subsec:disease_cohorts}. Sensitivity analysis using empirical sampling and different subsets of slides yielded consistent overall findings (Appendix~\ref{app:resampling}).

\subsubsection{Comparison to alternate ordering baselines}
To investigate whether simpler ordering procedures could recover comparable trajectory orderings from foundation model embeddings, we compared DPT against two baselines: (1) cosine distance from each patch to the centroid of the earliest-class embeddings, which assumes progression can be modeled as radial displacement from a starting point under approximately isotropic geometry; and (2) projection onto the first principal component of the embedding cohort, which assumes progression varies linearly along the dominant axis of variation. For each foundation model, DPT achieves higher mean trajectory fidelity across progressions than both baselines (14\% improvement over cosine distance from centroid, 36\% over PCA on average; Appendix~\ref{app:baselines}). We attribute this to DPT's weaker geometric assumptions: by relying on local smoothness rather than global isotropy or linearity, DPT can model progression manifolds with curvature and non-uniform sampling density.

To further test robustness to the choice of trajectory inference method, we compared DPT against two nonlinear alternatives that share its $k$-NN graph and root: the first diffusion component alone (DC1) and a $k$-NN-graph geodesic ordering. DPT retains the highest mean $\tau$ across progressions and models (0.64, vs.\ 0.60 for DC1 and 0.58 for the $k$-NN geodesic), and the orderings produced by the alternative methods agree closely with DPT at the patch level (Spearman $\rho = 0.76$--$0.86$; see Appendix~\ref{app:baselines}). Further, the ranking of models by average trajectory fidelity is largely stable across nonlinear inference methods, with UNI-2 and Virchow-2 always ranking above others and DINOv2 always ranking last (Appendix~\ref{app:baselines}). Together, these results suggest that DPT is an effective approach for recovering progression orderings from foundation model embeddings, but also that the orderings reflect the embedding structure itself rather than assumptions specific to DPT.

\subsubsection{Observed trajectory fidelity is not driven by class separability alone}
\label{subsubsec:permutation_test}

\begin{wrapfigure}{l}{0.5\columnwidth}
    \centering
    \includegraphics[width=0.48\columnwidth]{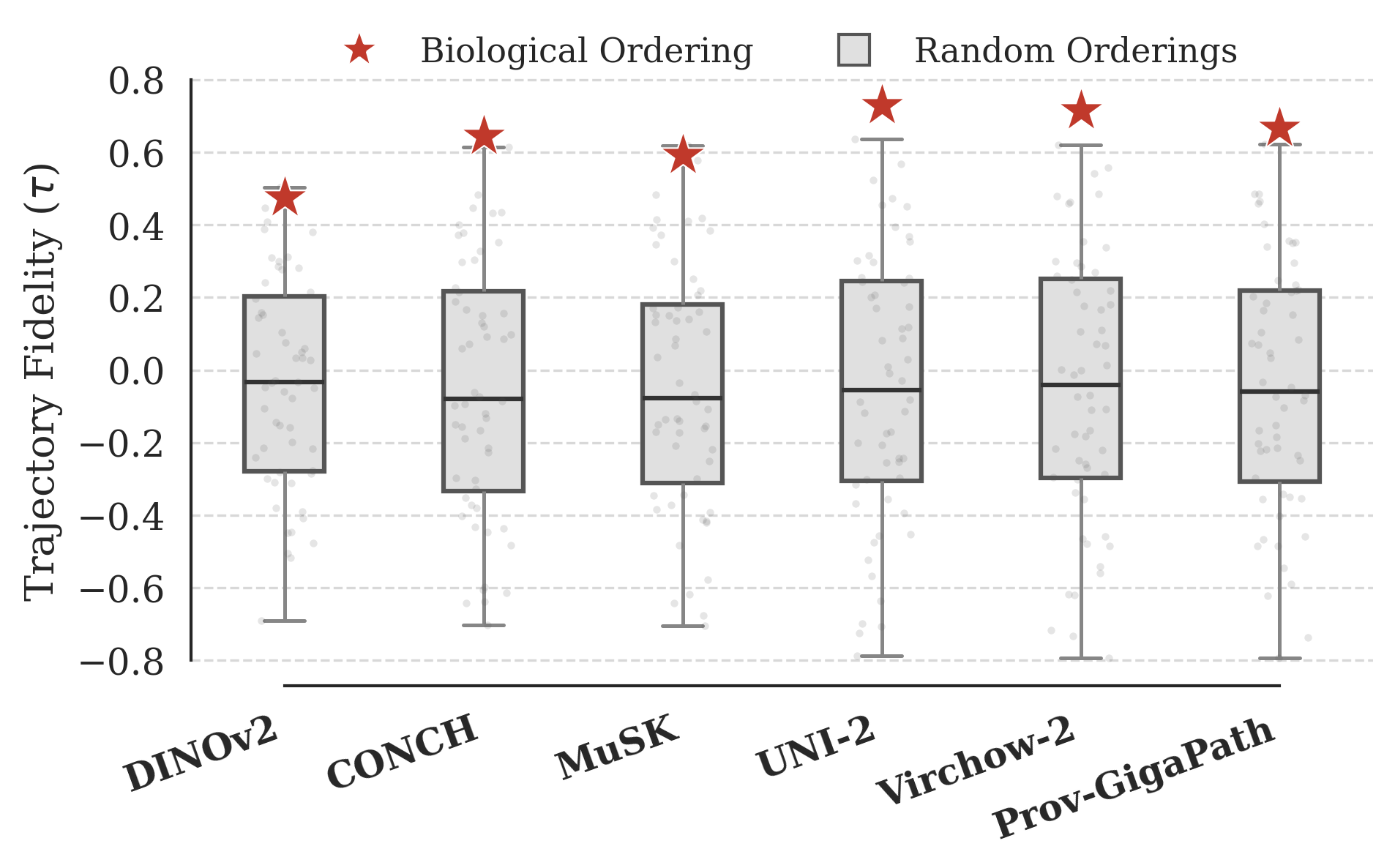}
    \caption{\textbf{Trajectory fidelity reflects biological ordering, not class separability alone.} For each model and progression, the inferred pseudotime is held fixed and $\tau$ is recomputed against every possible ordering of the class labels. Stars: mean $\tau$ for true progression order across four cohorts. Boxplots: distributions from permuted class label orderings, pooled across cohorts. }
    \label{fig:permutation_test}
    \vspace{-1\intextsep}
\end{wrapfigure}

We next analyzed whether the observed $\tau$ values were simply a result of class separability in embedding space, rather than reflecting an ordinal progression of classes along the manifold.
If the former were true (i.e., embeddings cluster by class and this clustering alone drives high $\tau$), then permuting the ground-truth ordering of classes in the $\tau$ calculation would lead to similar values to the observed value.
We thus computed $\tau$ for every possible ordinal sequence of the class labels, while keeping the inferred pseudotimes fixed.

We first examine results aggregated across all four disease progressions. For each model, we averaged the $\tau$ values from the true biological orderings across progressions and compared this to the distribution of $\tau$ values from all permuted orderings pooled across progressions.
As shown in Figure~\ref{fig:permutation_test}, the permuted distributions center around zero and for all CPath models, the true ordering mean $\tau$ falls outside or at the upper boundary of the permuted distribution, with vision-only models showing the clearest separation. This indicates that the embeddings do not merely separate disease states but arrange them in a geometry that specifically reflects the biological progression direction. Per-progression results confirm this pattern holds for individual pathologies (Appendix~\ref{app:permutation}), though with more variability, particularly for BDC with DINOv2 and MuSK.

\subsection{Emergence of Trajectory Fidelity Across Network Depth}
\label{subsec:emergence}

\begin{wrapfigure}{r}{0.5\columnwidth}
    \vspace{-\intextsep}
    \centering
    \includegraphics[width=0.46\columnwidth]{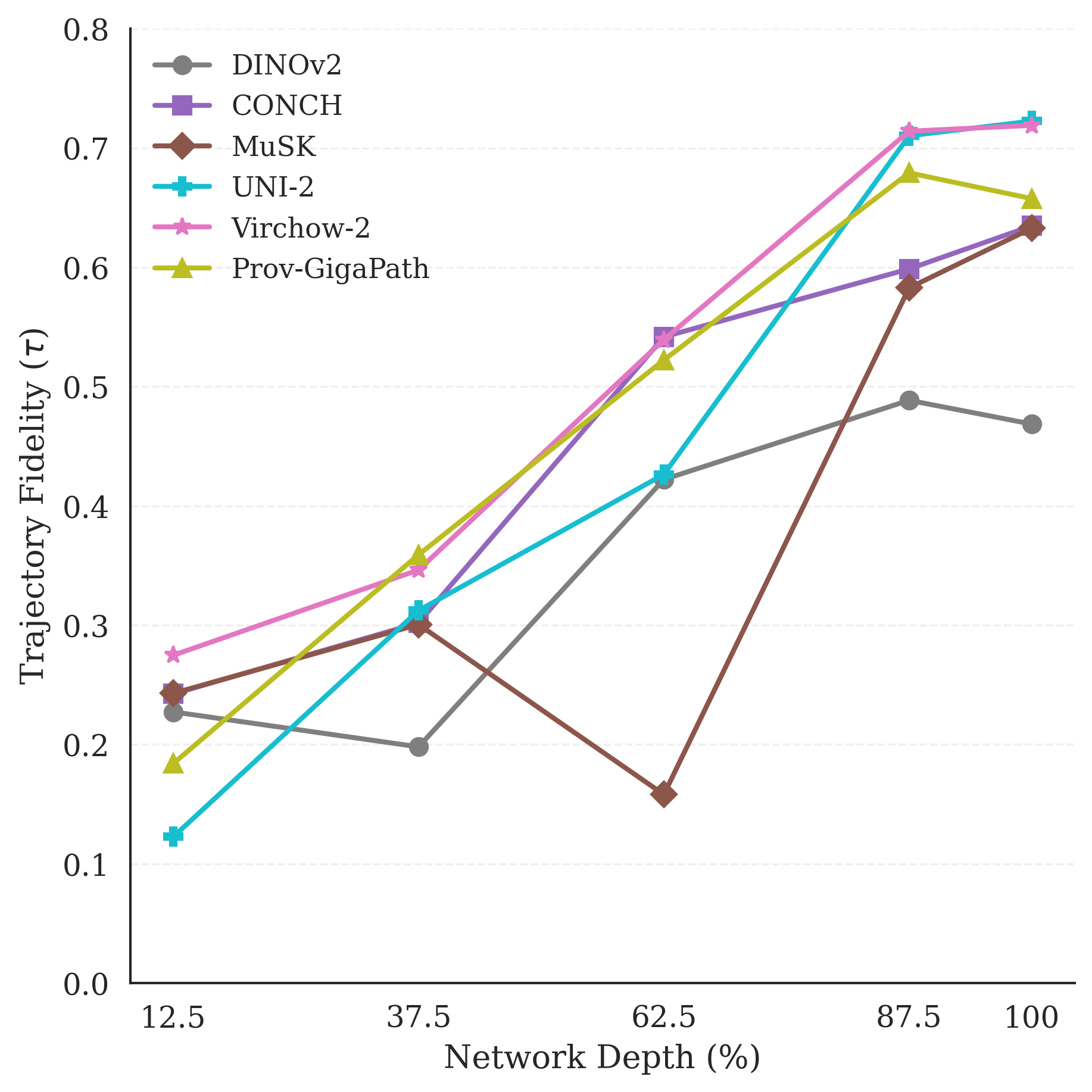}
    \caption{\textbf{Trajectory fidelity emerges progressively across network depth.} Trajectory fidelity ($\tau$) using [CLS] token embeddings, averaged across progressions. Vision-only pathology models (UNI-2, Virchow-2, Prov-GigaPath) and CONCH show steady improvement through 87.5--100\% depth, while MuSK exhibits a mid-network dip before recovery.}
    \label{fig:emergence}
    \vspace{-2\intextsep}
\end{wrapfigure}

When does the observed progression structure emerge in foundation model representations? To investigate this, we extracted [CLS] embeddings from intermediate transformer blocks at five relative depths (12.5\%, 37.5\%, 62.5\%, 87.5\%, and 100\%) across all models and progressions. As shown in Figure~\ref{fig:emergence}, average trajectory fidelity increases across network depth for all models, though the rate and pattern of emergence varies by architecture.

Early layers encode some progression-relevant structure: at 12.5\% depth, trajectory fidelity ranges from $\tau \approx 0.1$ to $\tau \approx 0.3$ across models (Figure~\ref{fig:emergence}). This structure is substantially refined through the network, with final values spreading to $\tau \approx 0.45$--$0.70$. The vision-only CPath models (Virchow-2, UNI-2, Prov-GigaPath) and CONCH show the strongest and most consistent improvement across layer depth. MuSK shows non-monotonic behavior, with trajectory fidelity declining from $\tau \approx 0.30$ at 37.5\% to $\tau \approx 0.16$ at 62.5\% before recovering to $\tau \approx 0.64$ at 100\%. This pattern may reflect its BEiT-3 architecture, which activates vision-language fusion experts only in later layers~\citep{wang2022imageforeignlanguagebeit}. At early layers, DINOv2 is consistent with the CPath models but ends with the lowest trajectory fidelity ($\tau < 0.5$ throughout), suggesting that while natural image pretraining can detect lower-level progression-relevant signals, domain-specific pretraining becomes increasingly important for refining this structure.

\subsection{Trajectory Fidelity Tracks Few-Shot Generalization}
\label{subsec:generalization}

We next asked whether trajectory fidelity correlates with downstream few-shot classification performance. These are geometrically distinct properties: trajectory fidelity measures preservation of biological ordering, while few-shot classification measures discriminative margins between class subspaces. While our analysis above suggests that the observed $\tau$ values are not driven by class separability alone, a model with higher $\tau$ may also show higher classification performance, but this is not guaranteed, and a correlation could suggest that encoding progression structure confers broader representational benefits.

We first examined the relationship between trajectory fidelity and classification performance within individual disease progressions (Appendix Figure~\ref{fig:within_disease}). For each progression, we computed the correlation between $\tau$ and 5-shot F1 score on label classification with a multi-class logistic regression model (see Appendix~\ref{app:within_disease} for methodological details). We find that these correlations are consistently strong: $\rho = 0.94$ (BDC), $\rho = 0.83$ (SCC), $\rho = 0.77$ (CRC-Conv), and $\rho = 0.77$ (CRC-Serr), with a mean of $\rho = 0.83$. This indicates that, controlling for disease-specific baseline difficulty, models that better preserve progression geometry achieve higher few-shot classification performance.

\begin{figure*}[t]
    \centering
    \includegraphics[width=\linewidth]{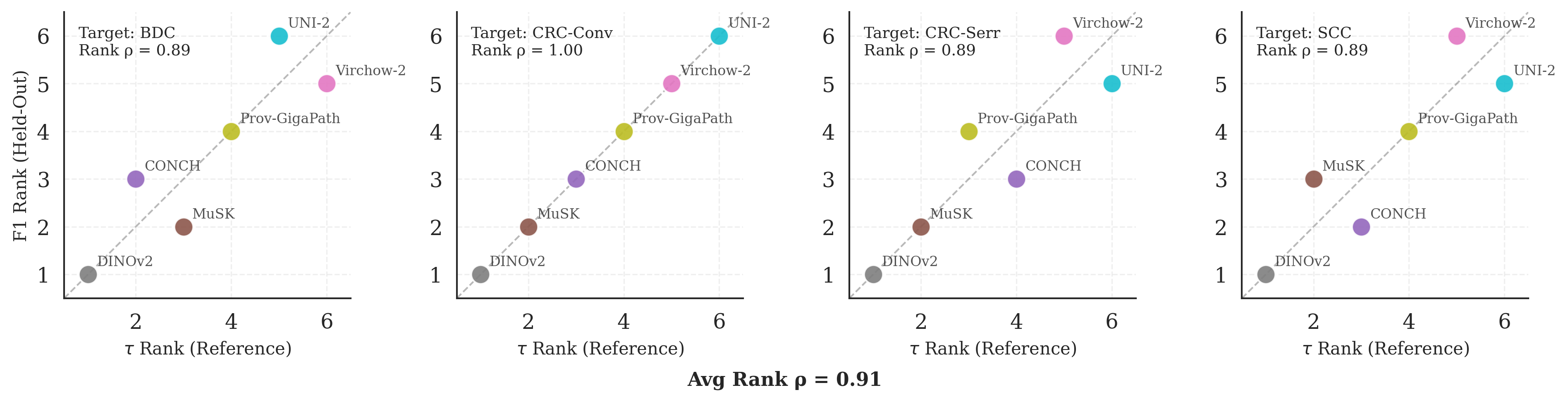}
    \caption{\textbf{Model rankings by trajectory fidelity transfer to few-shot classification of held-out diseases.} Cross-task generalization using a leave-one-out protocol. For each target disease, we ranked models by mean $\tau$ on the remaining cohorts (x-axis) and by 5-shot F1 on the held-out target (y-axis). Diagonal indicates perfect rank preservation. Model rankings transfer reliably across tissue types.}
    \label{fig:cross_task}
    \vspace{-\intextsep}
\end{figure*}

To test whether this relationship transfers across disease types, we employed a leave-one-out protocol. For each of the four progressions, we computed the mean trajectory fidelity across the three reference cohorts (excluding the target) and evaluated 5-shot classification F1 on the held-out progression. We then assessed whether model rankings by reference $\tau$ predict rankings by held-out F1 (Figure~\ref{fig:cross_task}). We observe consistently high Spearman rank correlations across all four held-out progressions: $\rho = 1.00$ (CRC-Conv), $\rho = 0.89$ (BDC), $\rho = 0.89$ (SCC), and $\rho = 0.89$ (CRC-Serr), with a mean of $\rho = 0.91$. This indicates that model rankings by trajectory fidelity on reference diseases reliably transfer to held-out progressions, suggesting that trajectory fidelity captures aspects of representation quality that generalize across disease types. Additional details on the experimental design are included in Appendix~\ref{appendix:cross_task}.

\subsection{Exploratory Analysis of Biological Features across Pseudotime}
\label{subsec:trajectory_exploration}

Having established that the embedding manifolds of CPath models encode disease trajectories, we explored what additional insights this affords. One important question in histopathology is how cellular composition changes throughout progression, potentially reflecting phenomena such as immune evasion. By providing an ordinal ranking of patches by pseudotime, our approach offers a more granular view of this composition than discrete class labels alone. To validate the biological relevance of this ordering, we analyzed the cellular composition of the colorectal conventional trajectory using the HistoPLUS cell segmentation and classification model \citep{adjadj2025comprehensivecellularcharacterisationhe}. HistoPLUS was trained on colon adenocarcinoma, specifically the TCGA-COAD dataset, and has demonstrated generalization to external cohorts. We quantified cell type abundance on $N=8{,}000$ patches evenly sampled from each class along pseudotime in the UNI-2 embedding space and visualized the standardized dynamics in Figure~\ref{fig:bio_validation}. Additional methodological details and results using Virchow-2 are contained in Appendix~\ref{app:histoplus_analysis}, showing similar findings.

\begin{wrapfigure}{l}{0.5\columnwidth}
    \vspace{-\intextsep}
    \centering
    \includegraphics[width=0.48\columnwidth]{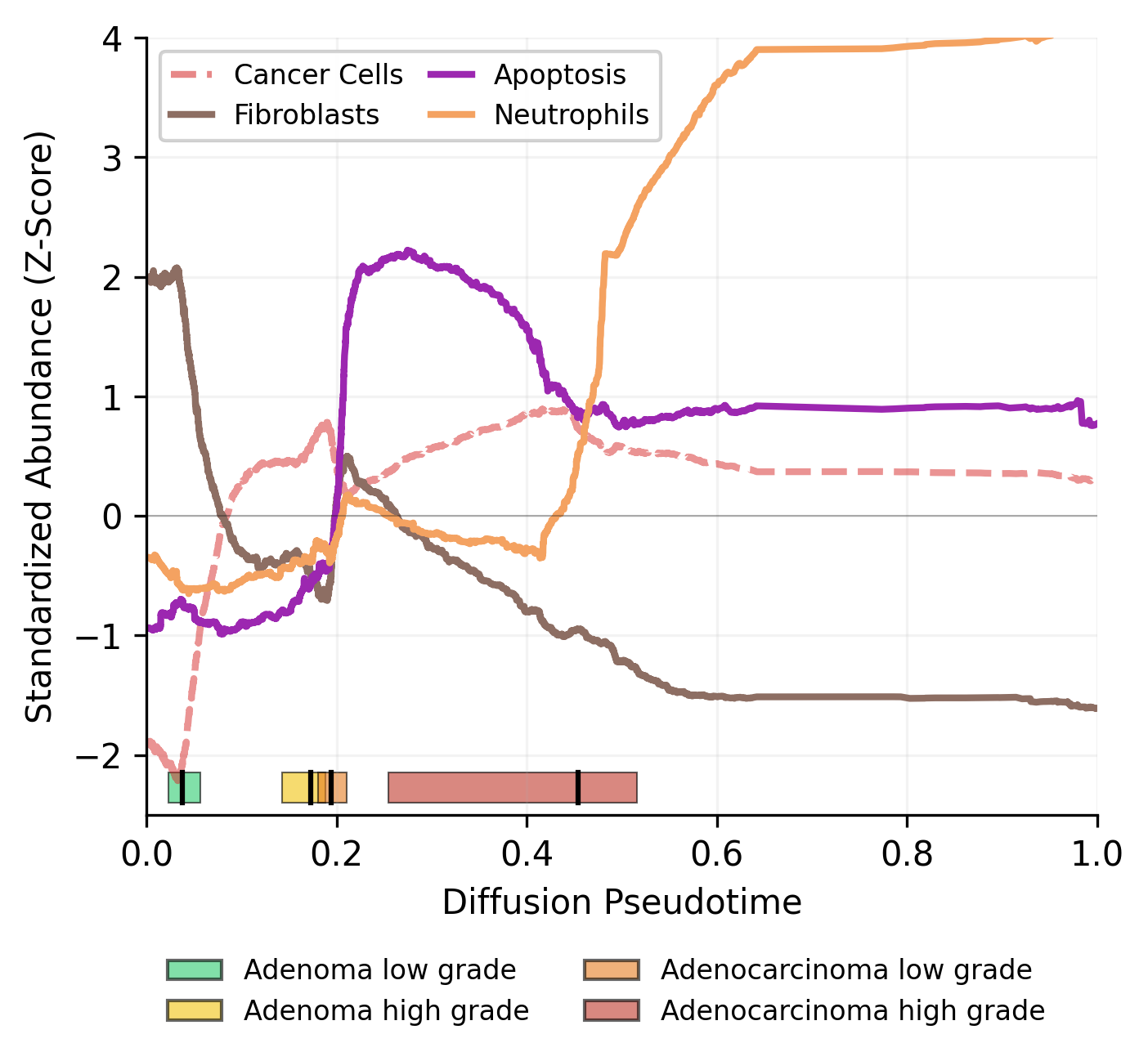}
    \caption{\textbf{Biological evaluation of the learned trajectory.} Patch-level cellular composition across UNI2-inferred pseudotime on the colorectal conventional progression cohort. Composition was quantified using HistoPLUS segmentation; mean proportions are Z-score standardized to compare trends across rare and dominant cell types. Ground-truth SPIDER class label distributions (edges: interquartile range; center line: median) shown above x-axis (see Appendix Fig.~\ref{fig:pseudotime_diagnostics} for full distributions).}
    \label{fig:bio_validation}
    \vspace{-\intextsep}
\end{wrapfigure}

\textbf{Consistency with the adenoma--carcinoma continuum.}
The inferred trajectory places high-grade adenoma and low-grade adenocarcinoma patches in close proximity in pseudotime, with overlapping interquartile ranges. This distribution matches prior pathology consensus that these classes can be difficult to separate visually when invasion is not captured in the local field of view \citep{schlemper2000vienna}.

\textbf{Changes in cell abundances across pseudotime may align with aspects of stromal remodeling.}
We observe a V-shaped pattern in fibroblast abundance centered near the adenoma--adenocarcinoma transition region.
Fibroblast abundance appears to decrease across the pseudotime region representing Adenoma high grade, starting from a z-scored value of around 2 at early pseudotime before reaching a local minimum of -0.7 at pseudotime 0.18. This result is consistent with epithelial overgrowth and back-to-back glandular architecture that compresses the stromal compartment, reducing the space occupied by fibroblasts during dysplasia \citep{Fleming2012}.
Subsequently, fibroblast abundance shows an increase over pseudotime for Adenocarcinoma low grade, reaching a z-scored value of 0.4 at pseudotime 0.21. This increase potentially reflects the onset of the desmoplastic reaction and activation of cancer-associated fibroblasts that accompany early invasive growth in colorectal carcinoma \citep{Ueno2021,Kalluri2016}.
This trend emerges smoothly over pseudotime despite substantial overlap in discrete class labels, suggesting that the inferred ordering captures meaningful structure that is obscured by coarse categorical annotations. See Appendix~\ref{app:pseudotime_diagnostics} for robustness analyses.

\textbf{Progressive ordering within advanced disease.}
Although cancer cell abundance increases early along pseudotime and plateaus by mid trajectory $(t \simeq 0.4)$, neutrophilic infiltration continues to increase through the ordered Adenocarcinoma high grade samples, consistent with the prognostic significance of intratumoral neutrophil density in colorectal cancer \citep{Shaul2019}.

\textbf{Non-morphological progression axes.} To explore whether the inferred orderings also track progression axes not explicitly defined by morphology, we applied the framework to pathological stage, aneuploidy score, and disease-specific survival in TCGA-KIRC; all CPath models exceed a label-shuffled null on each endpoint (Appendix~\ref{app:tcga}).

\section{Discussion}
\label{sec:discussion}

\textbf{Disease progression emerges as a low-dimensional ordered structure in CPath foundation model representations, with quantitative variations between models.} All evaluated CPath models recovered trajectory orderings significantly exceeding null baselines, demonstrating that their self-supervised learned embeddings organize disease states along coherent progression directions. Nonetheless, fidelity varied quantitatively across models and disease types: UNI-2 and Virchow-2 achieved the highest average fidelities and consistently ranked highly across progressions, while MuSK showed the lowest average fidelity and the greatest variability across cohorts; all CPath models exceeded the natural image DINOv2 baseline in average fidelity.

\textbf{Trajectory fidelity complements classification benchmarks by offering insights into ordinal progressions rather than discrete labels and classification performance.} Classification performance reflects how well a representation supports separation between labeled classes, not necessarily whether the representation encodes biological relationships between disease states.
In contrast, trajectory fidelity probes a stronger geometric property: not merely that classes are separable, but that their arrangement in representation space reflects the underlying biological progression class ordering. Our biological grounding analysis provides support for this distinction. The observed variation in fibroblast and neutrophil abundance across inferred pseudotime aligns with known patterns of stromal remodeling during CRC progression. These dynamics emerge within and across discrete class boundaries, providing validation for our approach and highlighting the potential of using pseudotime-ordered samples to investigate other progression processes.  
This level of validation and exploration is not accessible through classification metrics, which are agnostic to inter-class organization.

\textbf{Trajectory fidelity also offers a potential signal of generalization.} While some morphological features may be cancer type-specific, others may be shared across cancer types and reflect general tissue biology and disease progression (e.g., cell types, immune infiltration, stromal remodeling, necrosis). As foundation models are trained on large heterogeneous datasets, a model with a representation space that organizes a specific progression in a coherent manner (resulting in a high trajectory fidelity) may therefore better encode these generalizable processes and may thus also show high fidelity or classification performance on another cancer type. Consistent with this, 
we find that model rankings by $\tau$ on reference diseases track rankings by few-shot F1 (mean $\rho = 0.91$), indicating that trajectory fidelity reflects transferable representation quality. Together, these results position $\tau$ as a complementary, geometry-based criterion for evaluating representations, offering insight into progressive biological structure while providing a potential proxy for downstream generalization.

\paragraph{Limitations.}
\label{sub:limitations}
Several methodological constraints bound the scope of these findings. Our analysis relies on ergodicity assumptions common to snapshot-based studies: diffusion-based methods can recover relative topological ordering but cannot identify absolute temporal rates. Disease states are defined by coarse ordinal labels; although SPIDER provides patch-level expert annotations with surrounding context, tumor heterogeneity and limited field of view mean patches may lack diagnostic hallmarks of their assigned class. We evaluated a set of popular CPath models and a natural image baseline; the observed variations should be interpreted as qualitative trends rather than statistically supported claims about specific design choices. Successful pseudotime recovery demonstrates that the embedding geometry aligns with progression, but does not reveal which specific features drive this alignment or whether the model has learned causal mechanisms versus correlates. Our exploratory biological analysis utilizes a single segmentation model and cohort, so errors in cell classification would propagate into the inferred trajectories. Future work could also extend this trajectory framework to study branching or heterogeneous processes, as DPT can be used to model multiple trajectory endpoints \citep{haghverdi2016diffusion}. Finally, while we leveraged four cohorts from a multi-site, multi-scanner dataset with detailed annotations, it will be important in future work to further assess generalization, including explicitly quantifying robustness to stain and acquisition shift.

\paragraph{Conclusion.}
Our results demonstrate that CPath foundation models can represent progressive biological structure beyond what discrete classification benchmarks assess. This opens the door to progression-level analysis directly in the representation space, including probing disease biology at finer granularity than current classification systems permit. More broadly, the approach may extend to any vision domain where continuous processes are observed through static snapshots, offering a framework to study whether representations capture the dynamics underlying their training data.

\begin{ack}
W.L. acknowledges funding support from the Ellison Foundation, the Wong Family Award, the Louis B. Mayer Foundation, the National Institute of Biomedical Imaging and Bioengineering award R21EB035247, and the National Library of Medicine award R01LM014775.
\end{ack}

\begin{small}
  \bibliographystyle{unsrtnat}
  \bibliography{references}
\end{small}

\clearpage
\appendix
\section*{\LARGE Appendix}
\section{DPT Hyperparameter Selection and Sensitivity Analysis}
\label{app:hyperparameters}

\begin{figure}[htbp]
    \centering
    \includegraphics[width=\textwidth]{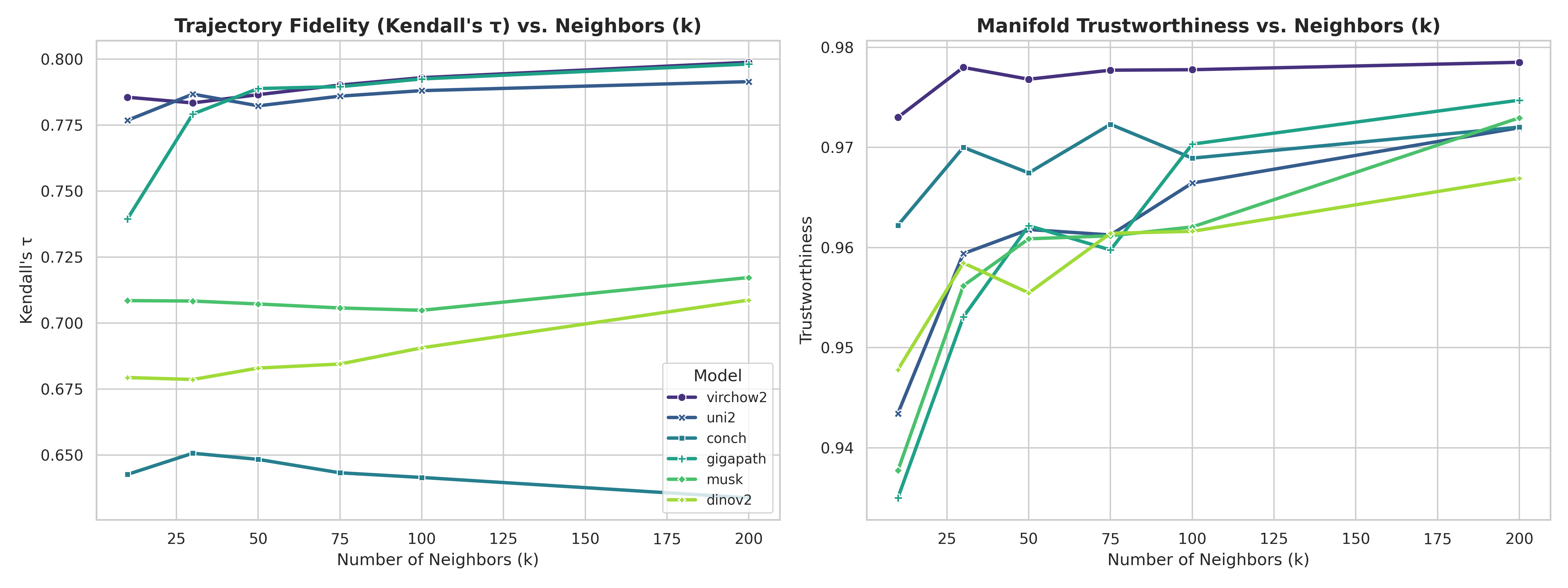}
    \caption{\textbf{Sensitivity analysis of the neighborhood parameter $k$.} We evaluated the impact of the number of nearest neighbors ($k$) on trajectory fidelity (Kendall's $\tau$) and manifold trustworthiness using the CRC-Serrated cohort. Trajectory fidelity (left) exhibits a stable plateau for $k \geq 50$ across leading pathology models, while trustworthiness (right) suggests that $k=100$ effectively preserves the local structure of the high-dimensional embeddings during diffusion mapping.}
    \label{fig:k_sweep}
\end{figure}

To apply the diffusion pseudotime method, the algorithm requires setting several hyperparameters: 
\begin{itemize}
\item \textbf{Root Selection.} Diffusion pseudotime requires designating a root sample as time zero. We use the medoid of the earliest disease class, which approximates the biological starting point without requiring manual selection. Sensitivity to root choice is limited: even if the selected root happens to represent an advanced case within the earliest class, a well-organized embedding space will still recover the correct global ordering, though early-class samples may exhibit increased pseudotime variance.

\item \textbf{Density Norm ($\alpha$):} The method uses Scanpy's default diffusion map, which applies the $\alpha$=1 density normalization.

\item \textbf{Diffusion Components:} We fixed the number of diffusion components at $n_{dc} = 10$ for all experiments. This choice was informed by intrinsic dimension (ID) analysis (Appendix~\ref{app:token_types}), which estimated the manifold dimensionality to be $\approx 4.5$ across all pathology-specific models. We chose a component count that exceeds the estimated ID by a factor of two to ensure complete capture of the biological progression signal while simultaneously acting as a spectral low-pass filter to exclude high-frequency noise and technical artifacts.
\item \textbf{K Nearest Neighbors:} We selected $k=100$ as the primary neighborhood size based on a hyperparameter sweep ($k \in [10, 200]$) evaluating Trajectory Fidelity and Manifold Trustworthiness. As shown in Figure \ref{fig:k_sweep}, trajectory fidelity remains stable for $k \geq 50$ across all high-performing pathology models. While Trustworthiness continues to show marginal gains up to $k=200$, the $k=100$ threshold represents a point of diminishing returns where the manifold structure is sufficiently captured without incurring the quadratic computational cost associated with larger neighborhood graph construction and multiscale diffusion distance calculations.

\end{itemize}

\FloatBarrier
\section{Slide-Aware Sampling Details}
\label{app:sampling}

To ensure trajectory fidelity reflects generalizable model properties rather than slide-level batch effects, we implemented a slide-aware sampling procedure. Naive random sampling would oversample densely-tiled WSIs while undersampling sparse ones, confounding trajectory inference with intra-slide correlations. For each disease progression cohort, we sample $n$ patches per class while enforcing a maximum of $m$ patches per slide:

For each class in the progression, we group patches by their source slide and shuffle the slide order. We then iterate through slides, taking up to $m_{\text{max\_per\_slide}}$ patches from each (shuffled within-slide) until reaching $n_{\text{per\_class}}$ total patches for that class.

The choice of $m_{\text{max\_per\_slide}} = 50$ balances within-slide morphological coverage against preventing any single slide from dominating trajectory inference. Combined with diffusion pseudotime's $k$-nearest-neighbor graph ($k = 100$), this reduces the likelihood that neighbor relationships are dominated by patches sharing preparation artifacts or scanner characteristics. The number of patches per class was set to 1,000 due to dataset size, and the desire to sample from multiple slides. 

\subsection{Sensitivity to slide selection and class balance}
\label{app:resampling}
 
The main results use class-balanced cohorts as described above and in Section~\ref{subsec:disease_cohorts}. To assess the stability of trajectory fidelity to slide selection and to the distribution of samples per class, we ran two resampling experiments on all four progressions (Table~\ref{tab:resampling}).
 
\paragraph{Slide selection.} We partitioned each cohort at the slide level into two disjoint halves (729/730 slides for SCC, 334/334 for CRC-Conventional, 343/343 for CRC-Serrated, 93/94 for BDC) and repeated balanced patch sampling and DPT within each half. Averaged across progressions, $\tau$ values and model rankings are similar across both halves and the full cohort: UNI-2 and Virchow-2 rank as the top two and DINOv2 ranks last in every condition. Most of the between-half variation arises in BDC, which contains fewer slides.
 
\paragraph{Empirical vs.\ balanced class distribution.} Because the diffusion normalization is density-dependent, we also recomputed $\tau$ after sampling patches according to the empirical distribution of classes within each disease, keeping the total patch count equal to the balanced cohort. Average $\tau$ values and rankings are again consistent, with the same top-two and bottom models under both distributions.
 
\begin{table}[h]
\centering
\caption{\textbf{Trajectory fidelity ($\tau$) under slide-level and class-distribution resampling.} Partitions A and B: disjoint slide-level halves of each cohort, each resampled to the balanced design (4{,}000 patches for SCC and CRC-Conventional, 3{,}000 for CRC-Serrated, 1{,}500 for BDC). Full: the balanced main-text cohort (Table~\ref{tab:trajectory_fidelity}; 2{,}714 patches for BDC, larger than its partitions). Empirical: patches sampled according to the observed class distribution with the same total count as Full.}
\label{tab:resampling}
\small
\begin{tabular}{llcccc}
\toprule
Progression & Model & Partition A & Partition B & Full (balanced) & Empirical \\
\midrule
SCC
  & DINOv2        & 0.493 & 0.495 & 0.476 & 0.495 \\
  & CONCH         & 0.621 & 0.633 & 0.637 & 0.641 \\
  & MuSK          & 0.558 & 0.602 & 0.641 & 0.641 \\
  & UNI-2         & 0.703 & 0.733 & 0.707 & 0.728 \\
  & Virchow-2     & 0.718 & 0.783 & 0.717 & 0.728 \\
  & Prov-GigaPath & 0.504 & 0.635 & 0.590 & 0.567 \\
\midrule
CRC-Conv.
  & DINOv2        & 0.629 & 0.558 & 0.517 & 0.550 \\
  & CONCH         & 0.724 & 0.731 & 0.703 & 0.731 \\
  & MuSK          & 0.715 & 0.732 & 0.676 & 0.700 \\
  & UNI-2         & 0.720 & 0.772 & 0.725 & 0.758 \\
  & Virchow-2     & 0.713 & 0.759 & 0.733 & 0.744 \\
  & Prov-GigaPath & 0.736 & 0.747 & 0.736 & 0.752 \\
\midrule
CRC-Serr.
  & DINOv2        & 0.712 & 0.708 & 0.691 & 0.707 \\
  & CONCH         & 0.709 & 0.693 & 0.642 & 0.669 \\
  & MuSK          & 0.664 & 0.728 & 0.705 & 0.719 \\
  & UNI-2         & 0.783 & 0.793 & 0.788 & 0.786 \\
  & Virchow-2     & 0.788 & 0.796 & 0.793 & 0.791 \\
  & Prov-GigaPath & 0.792 & 0.795 & 0.792 & 0.788 \\
\midrule
BDC
  & DINOv2        & 0.304 & $-0.100$ & 0.217 & 0.366 \\
  & CONCH         & 0.616 & 0.558 & 0.598 & 0.660 \\
  & MuSK          & 0.430 & 0.594 & 0.344 & 0.597 \\
  & UNI-2         & 0.603 & 0.595 & 0.699 & 0.702 \\
  & Virchow-2     & 0.600 & 0.547 & 0.619 & 0.574 \\
  & Prov-GigaPath & 0.238 & $-0.005$ & 0.545 & 0.522 \\
\midrule
Mean
  & DINOv2        & 0.534 & 0.415 & 0.475 & 0.530 \\
  & CONCH         & 0.668 & 0.654 & 0.645 & 0.675 \\
  & MuSK          & 0.592 & 0.664 & 0.591 & 0.664 \\
  & UNI-2         & 0.702 & 0.723 & 0.730 & 0.743 \\
  & Virchow-2     & 0.705 & 0.721 & 0.715 & 0.709 \\
  & Prov-GigaPath & 0.568 & 0.543 & 0.666 & 0.657 \\
\bottomrule
\end{tabular}
\end{table}

\section{Class separability vs. trajectory fidelity: additional motivation and results}
\label{app:tau_vs_separability}
\begin{figure}[htbp]
    \centering
    \includegraphics[width=0.95\textwidth]{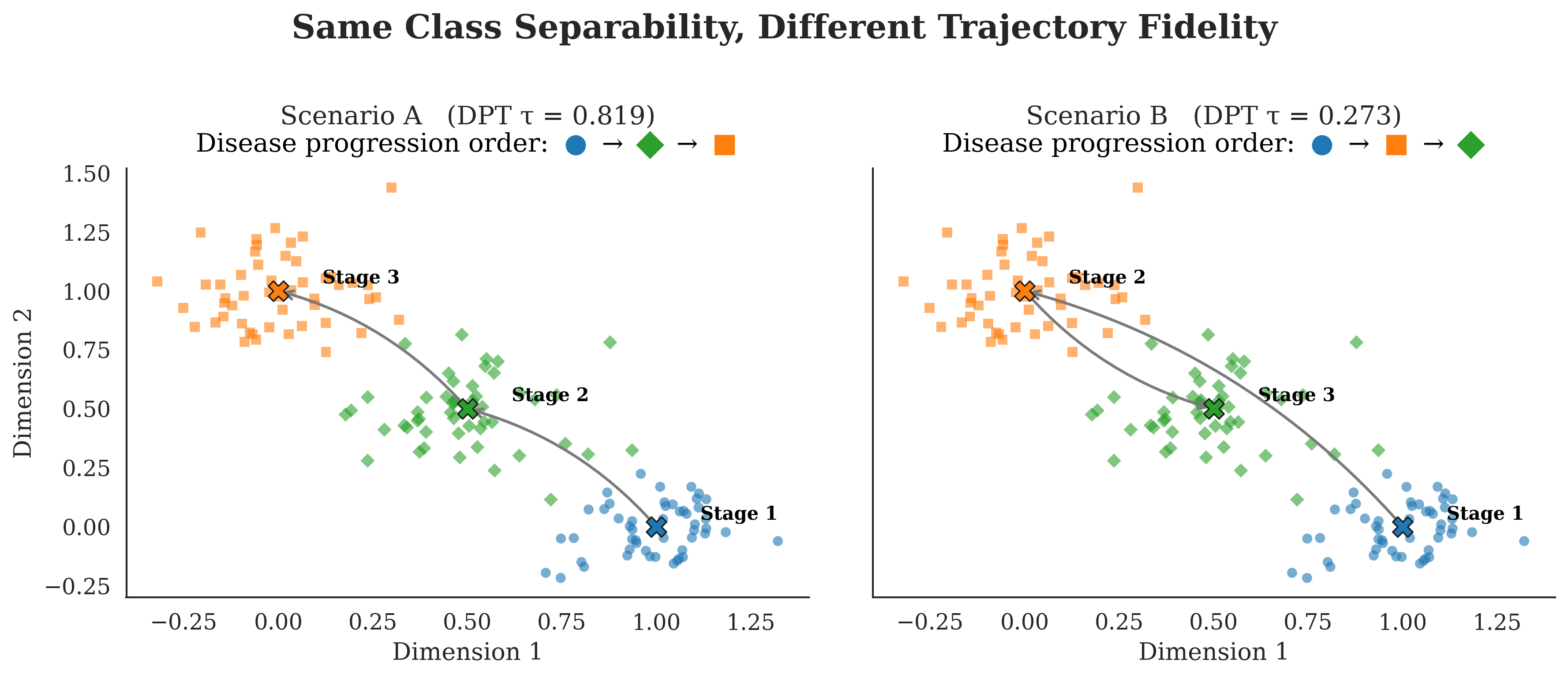}
    \caption{\textbf{Class separability does not imply trajectory fidelity.} Two simulated scenarios with identical cluster geometry but different stage-to-cluster assignments. \textit{Left (Scenario A):} stages are assigned such that the progression order traces a smooth path along the manifold, yielding high trajectory fidelity ($\tau = 0.819$). \textit{Right (Scenario B):} the same clusters are reassigned so the progression order zig-zags across the manifold, collapsing trajectory fidelity ($\tau = 0.273$) despite identical separability. Crosses mark class centroids; gray lines connect centroids in the assigned progression order.}
    \label{fig:tau_vs_separability}
\end{figure}

To illustrate that trajectory fidelity ($\tau$) imposes a stricter geometric criterion than class separability, we constructed a simple simulation in which two scenarios share identical class geometry but differ in how class clusters are arranged along the progression direction. We model a three-stage progression in a 2D embedding space, with class centroids at $(1, 0)$, $(0.5, 0.5)$, and $(0, 1)$. For each centroid, we generate 50 samples by adding isotropic Gaussian noise ($\sigma = 0.15$). In Scenario A, the centroids at $(1, 0)$, $(0.5, 0.5)$, and $(0, 1)$ correspond to Stages 1, 2, and 3 respectively, so the progression order traces a smooth path along the data manifold. In Scenario B, the Stage 2 and Stage 3 assignments are swapped while the cluster geometry is held fixed, so the progression order zig-zags across the manifold. By construction, class separability metrics (e.g., silhouette score, linear probe accuracy) are identical between the two scenarios. Applying diffusion pseudotime ($k = 15$ neighbors, $10$ diffusion components, Euclidean distance), we obtain $\tau = 0.819$ for Scenario A and $\tau = 0.273$ for Scenario B (Figure ~\ref{fig:tau_vs_separability}). This illustrates that $\tau$ depends not on whether stages are separable, but on whether their arrangement in representation space aligns with the biological progression direction.

\begin{figure}[htbp]
    \centering
    \includegraphics[width=0.85\textwidth]{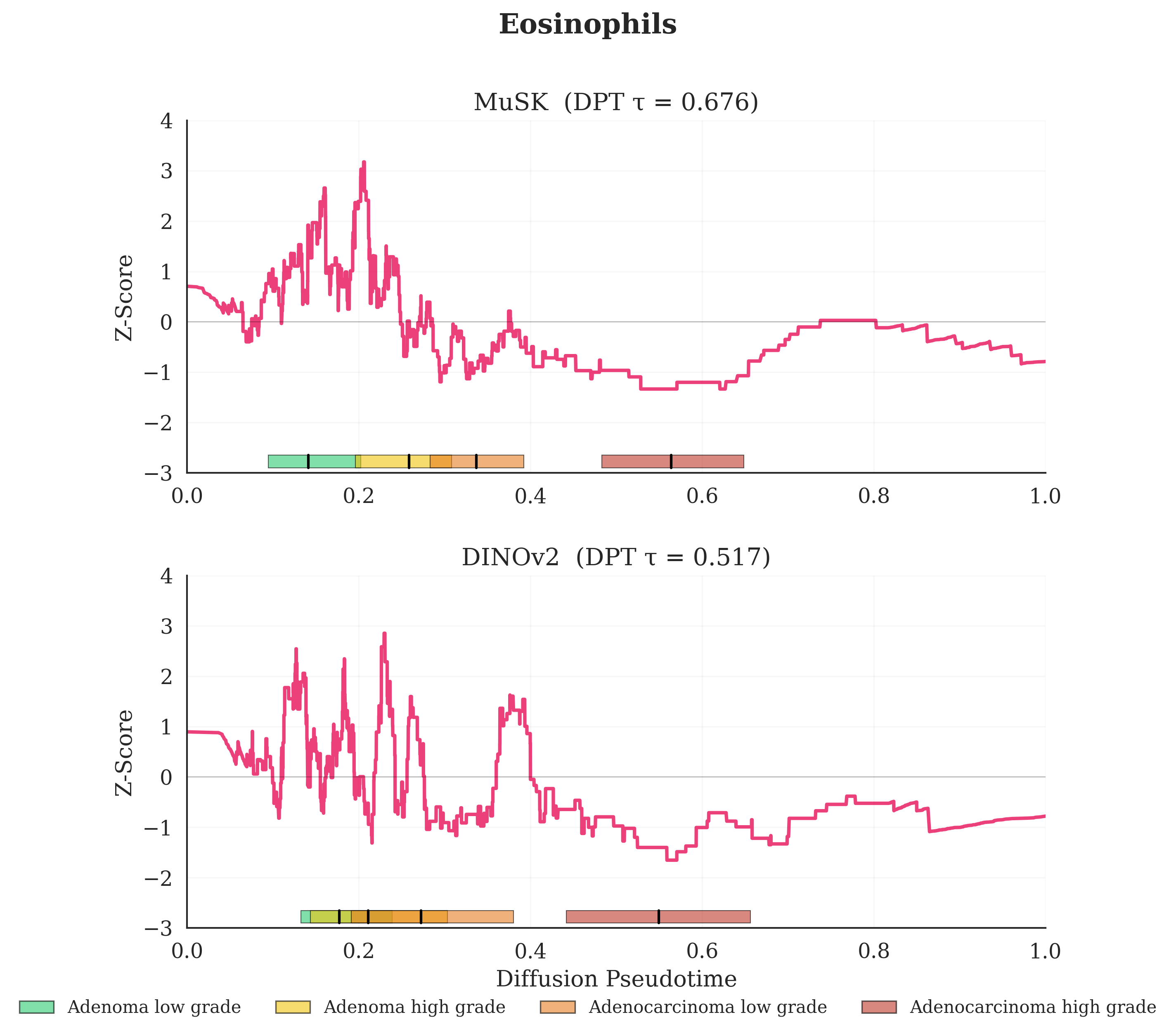}
    \caption{\textbf{Empirical example: similar class separability, different trajectory fidelity.} Eosinophil abundance (Z-score standardized) along pseudotime for the CRC-Conventional progression, computed using MuSK (top) and DINOv2 (bottom) embeddings.}
    \label{fig:musk_vs_dinov2_eosinophils}
\end{figure}

To ground this distinction empirically, we identified two models on CRC-Conventional, MuSK and DINOv2, that achieve nearly identical silhouette scores (0.062 average for both, indicating similar clustering geometry between classes) but differ in trajectory fidelity ($\tau = 0.68$ for MuSK vs.\ $\tau = 0.52$ for DINOv2). To assess whether this difference reflects meaningful structure, we examined cell-type dynamics along pseudotime as in Section~\ref{subsec:trajectory_exploration}. MuSK's ordering recovers an eosinophil pattern consistent with established pathology: peak infiltration during the adenoma classes followed by decline into adenocarcinoma \citep{moezzi2000stromal}. DINOv2's ordering shows greater oscillation, consistent with its lower $\tau$ reflecting interleaved samples across classes (Figure ~\ref{fig:musk_vs_dinov2_eosinophils}). Two models with comparable clustering metrics thus yield meaningfully different trajectory fidelity, and the difference aligns with known biology.

\FloatBarrier
\subsection{Comparison to alternate ordering baselines}
\label{app:baselines}

To assess whether the DPT formalism is necessary, we first compare against two simpler baselines: centroid distance, which uses cosine distance to the centroid of the earliest-class embeddings (assuming progression as radial displacement under approximately isotropic geometry), and PCA projection, which uses the first principal component coordinate of the cohort embeddings (assuming linear progression along the dominant variance axis). For each baseline, we compute Kendall's $\tau$ against ground-truth class labels using the same procedure as for DPT (Section~\ref{subsec:trajectory_fidelity}). The results are in Table~\ref{tab:baselines}.

Averaged across models and progressions, DPT exceeds centroid distance by 14\% and PCA projection by 36\% in mean percent improvement. 
Centroid distance tracks DPT closely on several pairs and occasionally exceeds it (e.g., CONCH on CRC-Serrated, Prov-GigaPath on SCC), but is consistently lower on average and degrades sharply on some model disease pairs. PCA projection collapses dramatically on a subset of pairs—most notably Prov-GigaPath on CRC-Conventional ($\tau = 0.22$ vs.\ DPT $\tau = 0.74$) and the top pathology models on CRC-Serrated ($\tau \approx 0.33$--$0.36$ vs.\ DPT $\tau > 0.78$), implying that progression-relevant variation is not always concentrated on the dominant variance axis. DPT's manifold-aware aggregation handles these cases more reliably without imposing global isotropy or linearity.

We additionally compare against two nonlinear alternatives that share DPT's $k$-nearest-neighbor graph ($k=100$, cosine distance) and root: the first diffusion component (DC1), which orders patches by the leading non-trivial eigenvector $\psi_1$ of $\tilde{\mathbf{T}}$ alone, and a $k$-NN geodesic ordering, which orders patches by shortest-path distance from the root over the $k$-NN graph. Averaged across progressions and models, DC1 and $k$-NN geodesic result in an average $\tau$ of 0.604 and 0.582, respectively, compared to 0.637 for DPT (Table~\ref{tab:baselines}).

\begin{table}[h]
\centering
\caption{\textbf{Trajectory fidelity ($\tau$) for DPT vs.\ alternative ordering baselines.} DPT achieves the highest mean $\tau$; the two nonlinear alternatives (DC1, $k$-NN geodesic) recover most of the same signal, while PCA projection collapses on several model-progression pairs (e.g., Prov-GigaPath on CRC-Conventional, top pathology models on CRC-Serrated). Bold marks the best method per row.}
\label{tab:baselines}
\small
\begin{tabular}{llccccc}
\toprule
Progression & Model & Centroid Dist. & PCA Proj. & $k$-NN Geod. & DC1 & DPT \\
\midrule
SCC
  & DINOv2        & 0.324 & 0.430 & 0.367 & 0.475 & \textbf{0.476} \\
  & CONCH         & 0.618 & 0.595 & 0.618 & 0.629 & \textbf{0.637} \\
  & MuSK          & 0.614 & 0.578 & 0.623 & 0.629 & \textbf{0.641} \\
  & UNI-2         & 0.668 & 0.686 & 0.673 & \textbf{0.721} & 0.707 \\
  & Virchow-2     & 0.622 & 0.653 & 0.647 & 0.714 & \textbf{0.717} \\
  & Prov-GigaPath & 0.645 & 0.636 & \textbf{0.659} & 0.586 & 0.590 \\
\midrule
CRC-Conv.
  & DINOv2        & 0.385 & 0.527 & 0.292 & \textbf{0.534} & 0.517 \\
  & CONCH         & 0.657 & 0.695 & 0.642 & \textbf{0.735} & 0.703 \\
  & MuSK          & 0.611 & 0.669 & 0.590 & \textbf{0.708} & 0.676 \\
  & UNI-2         & 0.707 & 0.716 & 0.713 & 0.683 & \textbf{0.725} \\
  & Virchow-2     & 0.704 & 0.640 & 0.732 & 0.700 & \textbf{0.733} \\
  & Prov-GigaPath & 0.611 & 0.223 & 0.661 & 0.677 & \textbf{0.736} \\
\midrule
CRC-Serr.
  & DINOv2        & 0.507 & 0.646 & 0.593 & \textbf{0.702} & 0.691 \\
  & CONCH         & \textbf{0.707} & 0.588 & 0.695 & 0.621 & 0.642 \\
  & MuSK          & 0.685 & 0.663 & 0.714 & \textbf{0.742} & 0.705 \\
  & UNI-2         & \textbf{0.790} & 0.351 & 0.786 & 0.736 & 0.788 \\
  & Virchow-2     & 0.794 & 0.332 & \textbf{0.796} & 0.658 & 0.793 \\
  & Prov-GigaPath & 0.732 & 0.362 & 0.758 & 0.544 & \textbf{0.792} \\
\midrule
BDC
  & DINOv2        & 0.117 & 0.192 & 0.044 & \textbf{0.341} & 0.217 \\
  & CONCH         & 0.469 & 0.549 & 0.407 & 0.516 & \textbf{0.598} \\
  & MuSK          & 0.381 & 0.407 & 0.241 & \textbf{0.456} & 0.344 \\
  & UNI-2         & 0.611 & 0.474 & 0.634 & 0.489 & \textbf{0.699} \\
  & Virchow-2     & \textbf{0.629} & 0.479 & 0.621 & 0.519 & 0.619 \\
  & Prov-GigaPath & 0.411 & 0.289 & 0.471 & 0.374 & \textbf{0.545} \\
\midrule
\textbf{Mean} & & 0.583 & 0.516 & 0.582 & 0.604 & \textbf{0.637} \\
\bottomrule
\end{tabular}
\end{table}

\begin{table}[t]
\centering
\small
\setlength{\tabcolsep}{5pt}
\caption{Model fidelity rankings under each ordering method. Within each row, models are listed from highest to lowest mean Kendall's $\tau$ across the four progressions; the mean $\tau$ is shown beneath each model.}
\label{tab:ordering_method_rankings}
\begin{tabular}{l*{6}{c}}
\toprule
& \multicolumn{6}{c}{Rank (1 = highest fidelity)} \\
\cmidrule(lr){2-7}
Ordering method & 1 & 2 & 3 & 4 & 5 & 6 \\
\midrule
DPT          & UNI-2 & Virchow-2 & GigaPath & CONCH & MuSK & DINOv2 \\
             & {\footnotesize 0.730} & {\footnotesize 0.715} & {\footnotesize 0.666} & {\footnotesize 0.645} & {\footnotesize 0.591} & {\footnotesize 0.475} \\
\addlinespace
kNN Geodesic & UNI-2 & Virchow-2 & GigaPath & CONCH & MuSK & DINOv2 \\
             & {\footnotesize 0.702} & {\footnotesize 0.699} & {\footnotesize 0.637} & {\footnotesize 0.590} & {\footnotesize 0.542} & {\footnotesize 0.324} \\
\addlinespace
Centroid     & UNI-2 & Virchow-2 & CONCH & GigaPath & MuSK & DINOv2 \\
             & {\footnotesize 0.694} & {\footnotesize 0.687} & {\footnotesize 0.613} & {\footnotesize 0.600} & {\footnotesize 0.573} & {\footnotesize 0.334} \\
\addlinespace
DC1          & UNI-2 & Virchow-2 & MuSK & CONCH & GigaPath & DINOv2 \\
             & {\footnotesize 0.657} & {\footnotesize 0.648} & {\footnotesize 0.634} & {\footnotesize 0.625} & {\footnotesize 0.545} & {\footnotesize 0.513} \\
\addlinespace
PCA Proj.    & CONCH & MuSK & UNI-2 & Virchow-2 & DINOv2 & GigaPath \\
             & {\footnotesize 0.607} & {\footnotesize 0.579} & {\footnotesize 0.557} & {\footnotesize 0.526} & {\footnotesize 0.449} & {\footnotesize 0.377} \\
\bottomrule
\end{tabular}
\end{table}

\paragraph{Distinguishing between ordering method and embedding quality.}
To further assess whether the recovered patch-level orderings belong to the embeddings or to DPT, we computed Spearman agreement between orderings for each pair of methods per model and progression. Averaged across models and progressions, DPT-based embedding orderings show a correlation of 0.86, 0.83, 0.76, and 0.68 to orderings derived from the $k$-NN geodesic, centroid distance, DC1, and PCA projection, respectively. The relatively high correlation between DPT-based embedding orderings and those from other non-linear methods thus suggests that the orderings are a reflection of the embedding structure itself rather than assumptions specific to DPT. 
Model fidelity rankings by mean $\tau$ across progressions (Table~\ref{tab:ordering_method_rankings})  are also stable across the nonlinear methods: DPT and the $k$-NN geodesic give identical orderings, centroid distance differs by one adjacent swap (CONCH and Prov-GigaPath), and DC1 differs in the positions of MuSK and Prov-GigaPath; in each case UNI-2 and Virchow-2 appear in the top two positions and DINOv2 appears last. Only the linear PCA projection produces a more distinct ranking, and the lowest mean $\tau$. These results are consistent with the disease progressions lying along nonlinear manifolds in the embedding space. 

\section{Trajectory fidelities for all models and progressions}
\label{app:traj_fidel_results}

In Table~\ref{tab:trajectory_fidelity}, we show all trajectory fidelities for all disease progressions and models, with the 95\% confidence intervals. 

\begin{table}[h]
\centering
\caption{Trajectory fidelity ($\tau$) scores across models and disease progressions. Values shown with 95\% confidence intervals.}
\label{tab:trajectory_fidelity}
\begin{tabular}{llccc}
\toprule
\textbf{Model} & \textbf{Progression} & \textbf{$\tau$} & \textbf{95\% CI} \\
\midrule
DINOv2 
 & BDC & 0.22 & [0.01, 0.28] \\
 & CRC-Conventional & 0.52 & [0.40, 0.59] \\
 & CRC-Serrated & 0.69 & [0.47, 0.71] \\
 & SCC & 0.48 & [0.12, 0.50] \\
\midrule
CONCH 
 & BDC & 0.60 & [0.53, 0.65] \\
 & CRC-Conventional & 0.70 & [0.52, 0.73] \\
 & CRC-Serrated & 0.64 & [0.62, 0.69] \\
 & SCC & 0.64 & [0.57, 0.66] \\
\midrule
MuSK 
 & BDC & 0.34 & [0.29, 0.54] \\
 & CRC-Conventional & 0.68 & [0.56, 0.71] \\
 & CRC-Serrated & 0.70 & [0.51, 0.71] \\
 & SCC & 0.64 & [0.44, 0.67] \\
\midrule
UNI-2 
 & BDC & 0.70 & [0.65, 0.72] \\
 & CRC-Conventional & 0.72 & [0.57, 0.74] \\
 & CRC-Serrated & 0.79 & [0.77, 0.79] \\
 & SCC & 0.71 & [0.49, 0.72] \\
\midrule
Virchow-2 
 & BDC & 0.62 & [0.56, 0.66] \\
 & CRC-Conventional & 0.73 & [0.72, 0.74] \\
 & CRC-Serrated & 0.79 & [0.79, 0.80] \\
 & SCC & 0.72 & [0.70, 0.73] \\
\midrule
Prov-GigaPath 
 & BDC & 0.54 & [0.45, 0.61] \\
 & CRC-Conventional & 0.74 & [0.66, 0.75] \\
 & CRC-Serrated & 0.79 & [0.75, 0.80] \\
 & SCC & 0.59 & [0.57, 0.62] \\
\bottomrule
\end{tabular}
\end{table}

\section{Manifold Robustness and Methodological Justification}
\label{app:manifold_robustness}

To verify that our manifold inference is robust to local variations in sampling density—particularly for the BDC cohort ($N=2{,}714$)—we evaluated the topological stability and dimensionality of the recovered representations.

\begin{figure}[htbp]
    \centering
    \includegraphics[width=\textwidth]{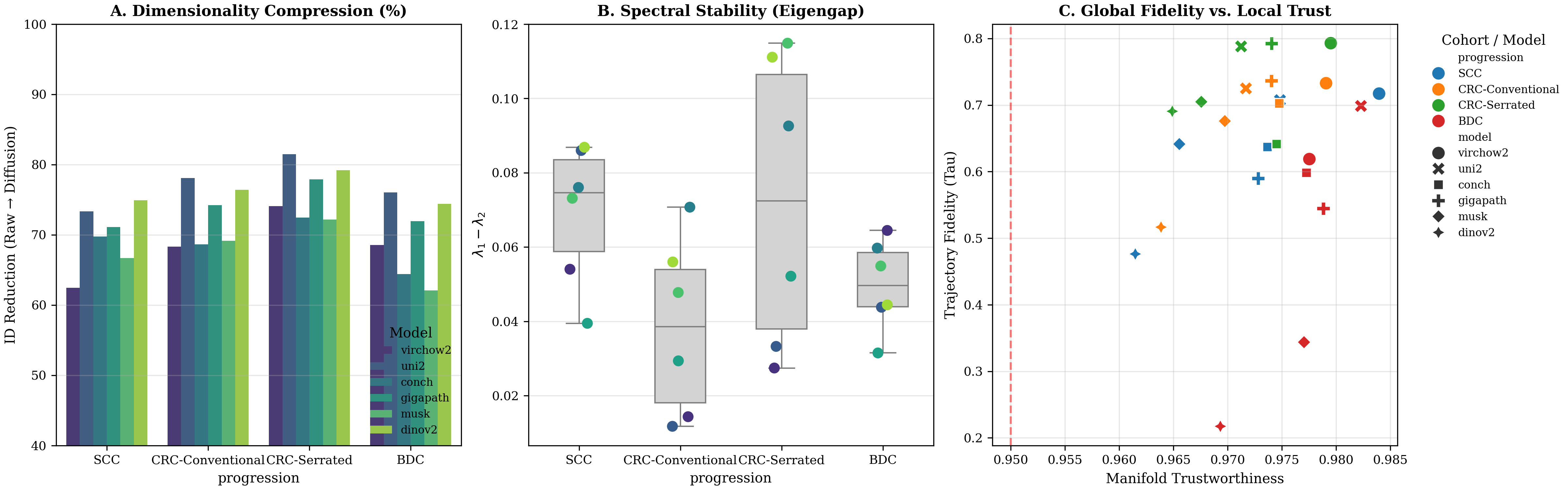}
    \caption{\textbf{Manifold Stability and Denoising Evidence.} (A) Dimensionality compression shows the percentage reduction from raw embedding ID to diffusion manifold ID. Values are $>60\%$ across all cohorts, suggesting successful denoising. (B) The spectral gap ($\lambda_1 - \lambda_2$) remains stable across all progressions; notably, the BDC cohort exhibits connectivity comparable to larger cohorts, supporting our slide-aware sampling strategy. (C) Global trajectory fidelity ($\tau$) is plotted against local manifold trustworthiness. High trustworthiness ($>0.95$) across all models ensures that the pseudotime trajectory is grounded in local morphological similarity without significant topological distortion.}
    \label{fig:manifold_robustness}
\end{figure}

\subsection{Intrinsic Dimension and Topological Denoising}

A core justification for using Diffusion Pseudotime (DPT) is its ability to perform topological denoising. As shown in Figure \ref{fig:manifold_robustness}A, we observe a consistent reduction in intrinsic dimensionality (ID) of approximately 70\% across all models and cohorts. For example, in the Virchow-2 BDC manifold, the ID is reduced from 12.7 to 4.0. This significant compression indicates that the diffusion operator successfully identifies a low-dimensional biological manifold from within the high-dimensional, noisy embedding space.

\subsection{Empirical Validation of the BDC Cohort}

Readers may note that the BDC cohort contains fewer patches ($N=2{,}714$) than other cohorts. However, our analysis suggests that the \textit{diversity} of the support is more critical than raw density. The BDC cohort exhibits a mean spectral gap of 0.050, which is higher than that of the CRC-Conventional cohort (0.038), despite the latter having 47\% more samples. Furthermore, BDC achieved the highest mean manifold trustworthiness (0.977) among all progressions. These results suggest that our slide-aware sampling (max 50 patches per slide) preserves the connectivity required for stable manifold inference even in data-constrained scenarios.

\begin{table}[htbp]
\centering
\caption{\textbf{Aggregated Manifold Quality Metrics.} Mean values calculated across all six foundation models. BDC shows similar values to the larger cohorts.}
\label{tab:manifold_metrics}
\begin{small}
\begin{tabular}{lcccc}
\toprule
\textbf{Progression} & \textbf{Spectral Gap} & \textbf{ID Reduction (\%)} & \textbf{Trustworthiness} & \textbf{Avg. $\tau$} \\
\midrule
BDC & 0.050 & 69.57\% & 0.977 & 0.504 \\
CRC-Conventional & 0.038 & 72.47\% & 0.972 & 0.682 \\
CRC-Serrated & 0.072 & 76.20\% & 0.972 & 0.735 \\
SCC & 0.069 & 69.70\% & 0.972 & 0.628 \\
\bottomrule
\end{tabular}
\end{small}
\end{table}

\FloatBarrier

\section{Stage Ordering Permutation Test Details}
\label{app:permutation}

\begin{figure}[htbp]
    \centering
    \includegraphics[width=\textwidth]{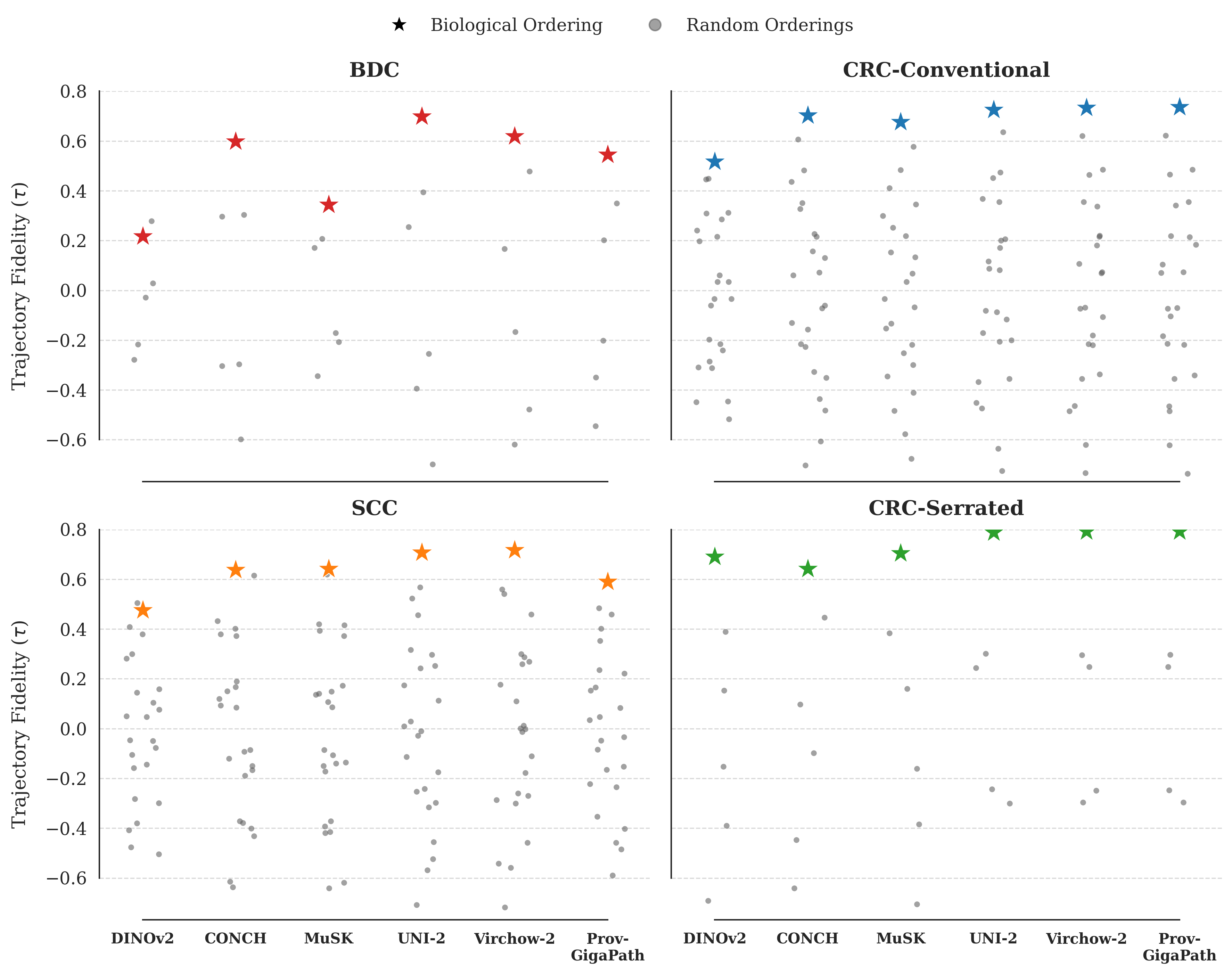}
    \caption{\textbf{Per-progression permutation test results.} Stars indicate $\tau$ for the true biological ordering; gray dots show results for all possible class order permutations. BDC exhibited the most model-dependent variation: DINOv2 and MuSK approached null performance while vision-only pathology models (UNI-2, Virchow-2, Prov-GigaPath) maintained stronger separation. Colorectal progressions (CRC-Conventional, CRC-Serrated) and SCC showed consistent separation from null across all models.}
    \label{fig:permutation_test_full}
\end{figure}

Figure~\ref{fig:permutation_test_full} presents the permutation test results stratified by disease progression. While the averaged results (Figure~\ref{fig:permutation_test}) demonstrate that all models encode progression structure beyond class separation, the per-progression view reveals cohort-specific variation. BDC proved most challenging: DINOv2 and MuSK yielded $\tau$ values within or near the null distribution. In contrast, the vision-only pathology models maintained separation from null across all four progressions, indicating generally more robust encoding of biological continuity.
\FloatBarrier
\section{Additional Layer Depth Analysis}
\label{app:token_types}

\begin{figure*}[htbp]
    \centering
    \includegraphics[width=\textwidth]{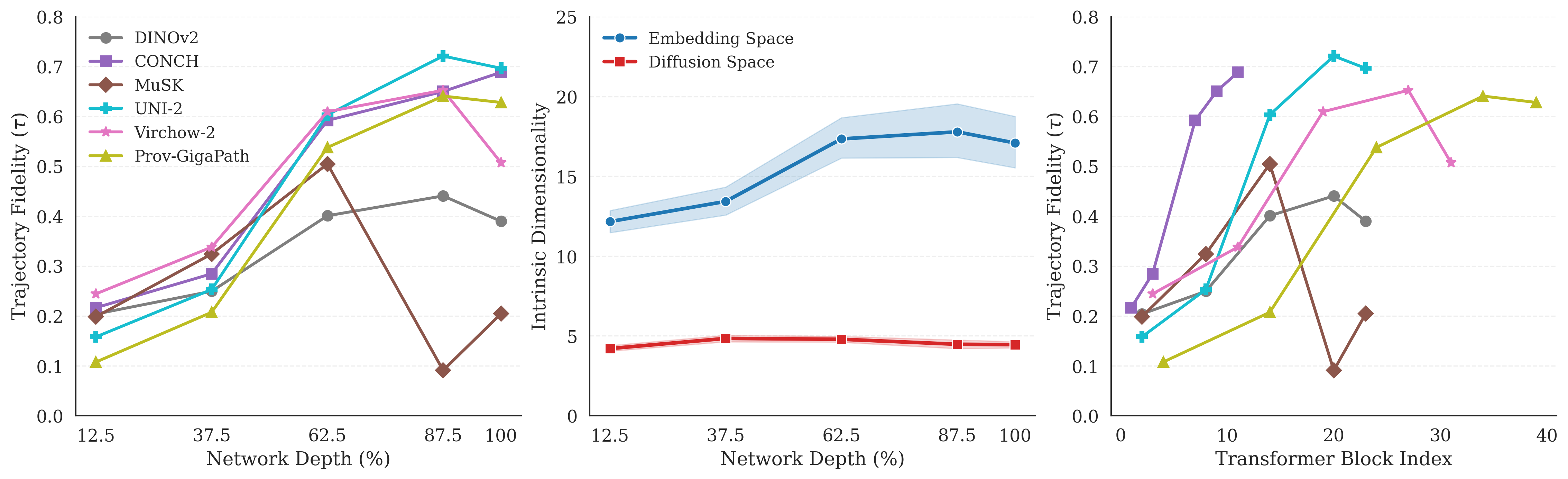}
    \caption{\textbf{Emergence analysis: Patch mean embeddings.} \textit{Left:} Trajectory fidelity ($\tau$) versus relative network depth. Steepest gains occur between 30--60\% depth; vision-only models continue improving through 87.5\%. MuSK collapses after 62.5\%. \textit{Center:} Raw embedding intrinsic dimensionality (blue) expands while diffusion manifold dimensionality (red, $\approx$4.5) remains stable. \textit{Right:} Absolute block index obscures the consistent relative-depth pattern due to varying model sizes (12--40 blocks).}
    \label{fig:emergence_patch_mean_full}
\end{figure*}

\begin{figure*}[htbp]
    \centering
    \includegraphics[width=\textwidth]{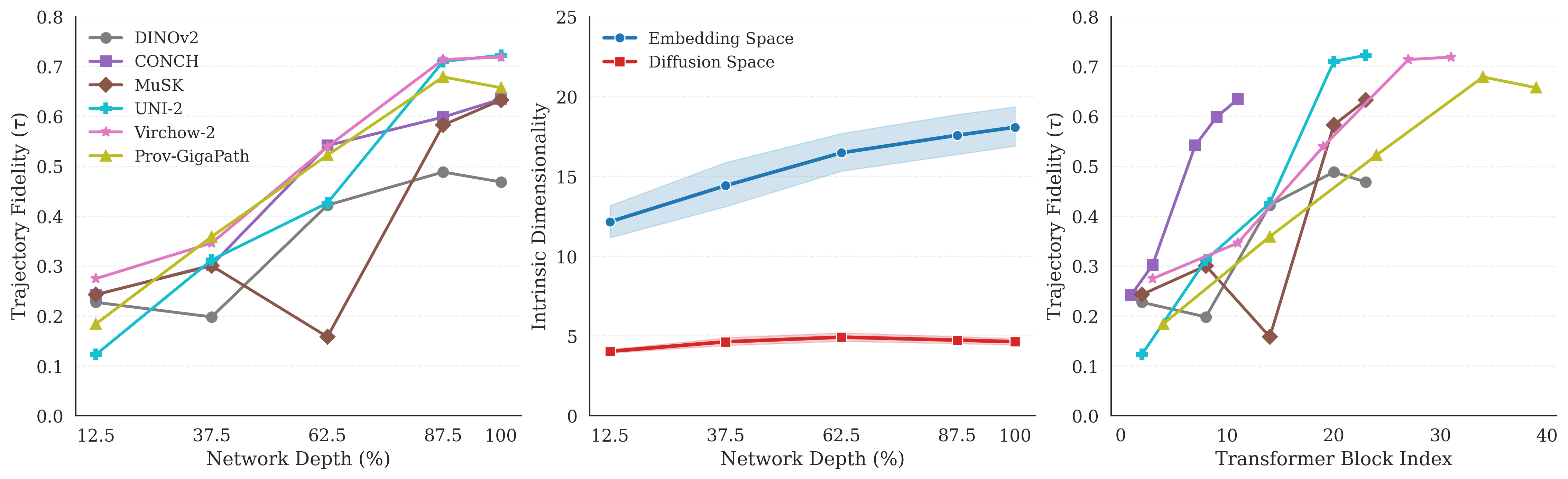}
    \caption{\textbf{Emergence analysis: CLS token embeddings.} \textit{Left:} CLS tokens show more monotonic trajectory fidelity increase compared to patch mean, particularly for vision-only models (UNI-2, Virchow-2, Prov-GigaPath) which improve steadily through final layers. MuSK exhibits the same mid-depth collapse as in patch mean, though less severe (recovering partially by 100\%). \textit{Center:} Intrinsic dimensionality patterns mirror those of patch mean. \textit{Right:} Absolute depth view.}
    \label{fig:emergence_cls}
\end{figure*}

\begin{figure*}[htbp]
    \centering
    \includegraphics[width=\textwidth]{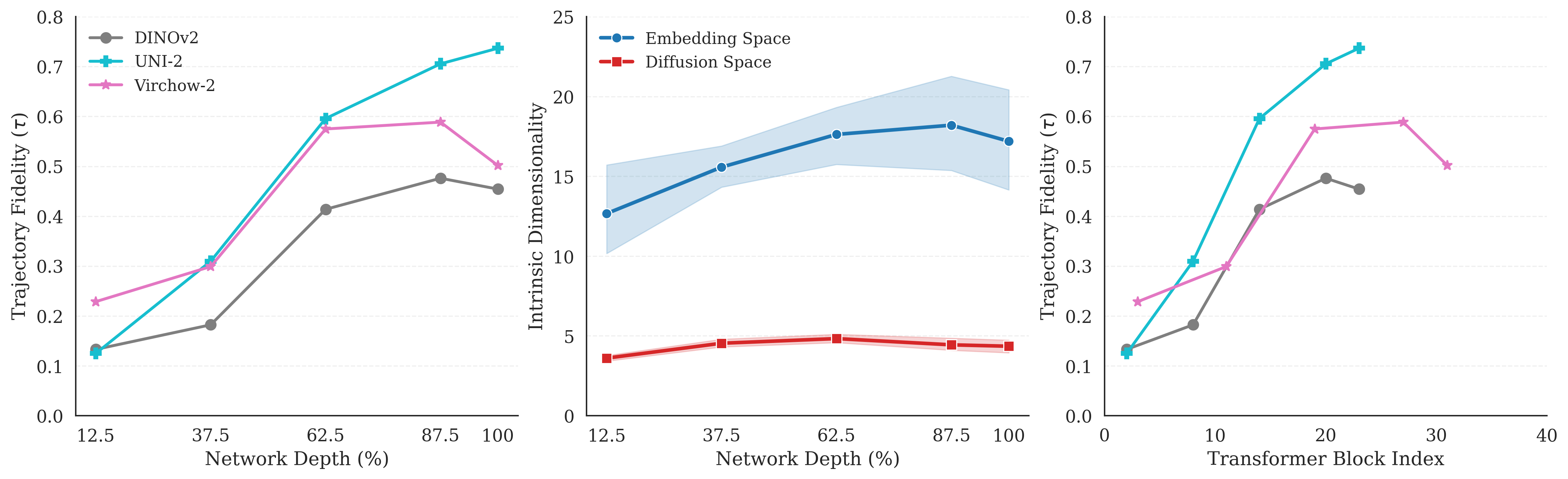}
    \caption{\textbf{Emergence analysis: Register mean embeddings.} Only models with register tokens are shown (DINOv2, UNI-2, Virchow-2). \textit{Left:} Register tokens show relatively consistent emergence curves. \textit{Center:} Intrinsic dimensionality patterns are consistent with other token types, though with wider confidence intervals due to fewer models. \textit{Right:} Absolute depth view confirms the relative-depth normalization aligns emergence across architectures.}
    \label{fig:emergence_register}
\end{figure*}

To further explore the emergence of trajectory fidelity by layer depth, we performed DPT analysis across three token types: patch mean, CLS, and register mean. Figures~\ref{fig:emergence_patch_mean_full}--\ref{fig:emergence_register} present three-panel analyses for each, showing trajectory fidelity versus relative depth, intrinsic dimensionality divergence, and trajectory fidelity versus absolute transformer block index. Intrinsic dimensionality (ID) quantifies the minimum number of variables needed to describe the representation's structure; a low ID relative to the raw embedding dimension indicates compression onto a lower-dimensional manifold. We use the TWO-NN estimator~\citep{facco2017estimating} applied to 2,000 subsampled patches.

When inspecting the [CLS] token in Figure~\ref{fig:emergence_cls}, we see early layers encode some progression-relevant structure: at 12.5\% depth, trajectory fidelity ranges from $\tau \approx 0.1$ to $\tau \approx 0.3$ across models (Figure~\ref{fig:emergence_cls}, Left). This structure is substantially refined through the network, with final values spreading to $\tau \approx 0.45$--$0.70$. The vision-only CPath models (Virchow-2, UNI-2, Prov-GigaPath) and CONCH show the strongest and most consistent improvement, with steady gains that steepen in later layers. MuSK shows non-monotonic behavior, with trajectory fidelity declining from $\tau \approx 0.30$ at 37.5\% to $\tau \approx 0.16$ at 62.5\% before recovering to $\tau \approx 0.64$ at 100\%. This pattern may reflect its BEiT-3 architecture, which activates vision-language fusion experts only in later layers~\citep{wang2022imageforeignlanguagebeit}. At early layers, DINOv2 is consistent with the CPath models but ends with the lowest trajectory fidelity ($\tau < 0.5$ throughout), suggesting that while natural image pretraining can detect lower-level progression-relevant signals, domain-specific pretraining becomes increasingly important for refining this structure. 

The refinement of trajectory fidelity coincides with expanding representational complexity (Figure~\ref{fig:emergence_cls}, Middle). Raw embedding intrinsic dimensionality increases from $\sim$12 to $\sim$18 dimensions across layers, tracking trajectory fidelity improvement. Yet diffusion manifold dimensionality remains stable at $\approx$4--5 dimensions throughout, suggesting that deeper layers build richer representational capacity while the underlying biological progression signal remains intrinsically low-dimensional. 

Patch token mean embeddings exhibit qualitatively similar but also differing emergence dynamics (Figure~\ref{fig:emergence_patch_mean_full}). Trajectory fidelity generally increases over network depth for most models, with the exception of MuSK. Diffusion space intrinsic dimensionality remains relatively stable across depth, similar to [CLS] tokens. However, embedding space intrinsic dimensionality follows a more sigmoidal trajectory, in contrast to the approximately linear increase observed for [CLS] tokens. 

\FloatBarrier

\section{Trajectory Fidelity and Classification Performance}
\label{app:gen}

This appendix provides methodological details for the analyses in Section~\ref{subsec:generalization}, which examine the relationship between trajectory fidelity ($\tau$) and few-shot classification performance.

\subsection{Within-Disease Correlation}
\label{app:within_disease}

We first assessed whether trajectory fidelity correlates with classification performance when both metrics are measured on the same disease progression.

\paragraph{Few-Shot Classification.}
We trained logistic regression probes with limited labeled examples to evaluate discriminative performance. For each trial, we sampled exactly $k=5$ patches per class uniformly at random (without replacement) for training, reserving all remaining patches for evaluation. Embeddings were z-score normalized using statistics computed solely from the training samples. We report macro-averaged F1 scores to account for class imbalance. To reduce sampling variance, we repeated each experiment across $N=10$ random trials and report mean performance.

\paragraph{Correlation Analysis.}
For each of the four disease progressions independently, we computed two metrics for each model:
\begin{enumerate}
    \item \textbf{Trajectory fidelity ($\tau$)}: Kendall's $\tau$ between diffusion pseudotime and ordinal class labels, calculated from final-layer embeddings as described in Section~\ref{sec:methods}.
    \item \textbf{5-shot F1}: Mean macro-averaged F1 score from the few-shot classification protocol above.
\end{enumerate}
We then computed Spearman's rank correlation ($\rho$) between these two metrics across the six models, yielding one correlation coefficient per disease progression.

\begin{figure}[htbp]
    \centering
    \includegraphics[width=\textwidth]{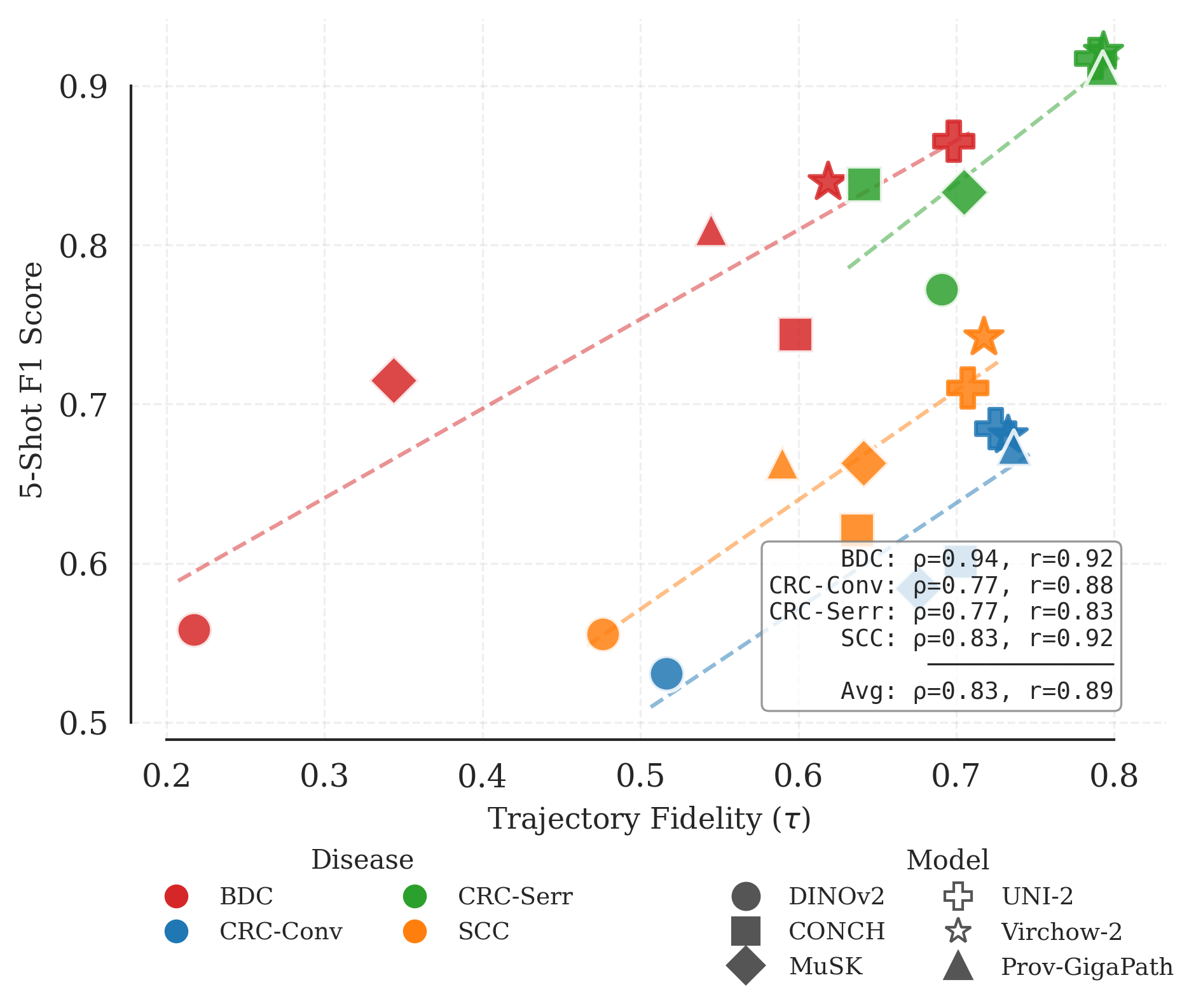}
    \caption{\textbf{Trajectory fidelity correlates with classification performance within each disease.} For each progression, we plot trajectory fidelity ($\tau$, x-axis) against 5-shot classification F1 (y-axis) for each model. Correlations are observed across all four diseases (mean $\rho = 0.83$), indicating that models preserving progression structure also generally achieve better discriminative performance.}
    \label{fig:within_disease}
\end{figure}

\paragraph{Results.}
Figure~\ref{fig:within_disease} shows consistently strong correlations: $\rho = 0.94$ (BDC), $\rho = 0.83$ (SCC), $\rho = 0.77$ (CRC-Conv), and $\rho = 0.77$ (CRC-Serr), with a mean of $\rho = 0.83$.

\FloatBarrier
\subsection{Cross-Task Generalization}
\label{appendix:cross_task}

We next tested whether trajectory fidelity on reference diseases predicts few-shot classification performance on held-out diseases, assessing whether $\tau$ captures transferable aspects of representation quality rather than disease-specific structure.

\paragraph{Leave-One-Out Protocol.}
We employed a leave-one-out cross-validation design across the four disease progressions. For each iteration:
\begin{enumerate}
    \item \textbf{Designate target}: One progression was held out as the target task.
    \item \textbf{Compute reference $\tau$}: For each model, we calculated the mean trajectory fidelity across the three remaining (reference) progressions.
    \item \textbf{Evaluate held-out F1}: We measured 5-shot classification F1 on the held-out target progression using the same few-shot protocol described above.
    \item \textbf{Correlate}: We computed Spearman's $\rho$ between reference $\tau$ and held-out F1 across the six models.
\end{enumerate}
This yields four correlation coefficients (one per target task), which we summarize by their mean.

\paragraph{Results.}
Figure~\ref{fig:cross_task} (main text) shows that reference trajectory fidelity predicts held-out classification performance: $\rho = 1.00$ (CRC-Conv), $\rho = 0.89$ (BDC), $\rho = 0.89$ (SCC), and $\rho = 0.89$ (CRC-Serr), with a mean of $\rho = 0.91$. This indicates that model rankings by trajectory fidelity on reference diseases reliably transfer to held-out tissue types.

\paragraph{Comparison with Classification as Reference Metric.}
To assess whether trajectory fidelity offers distinct predictive value, we compared it against classification performance as an alternative reference metric. Using the same leave-one-out protocol, we computed both reference $\tau$ (mean trajectory fidelity on three cohorts) and reference F1 (mean 5-shot classification F1 on three cohorts) for each model, then correlated each with held-out F1 on the fourth cohort.

\begin{figure}[htbp]
    \centering
    \includegraphics[width=\textwidth]{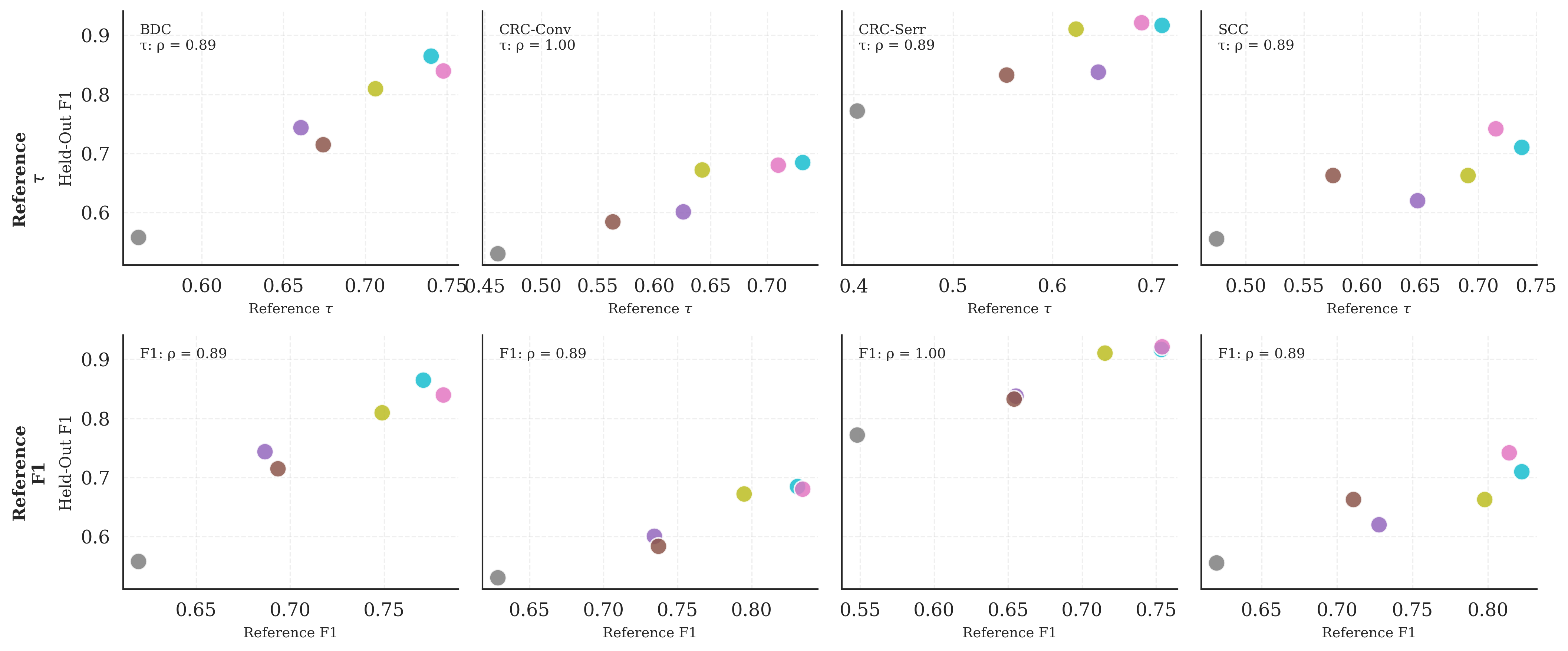}
    \caption{\textbf{Trajectory fidelity and classification performance are similar predictors of cross-task generalization.} We compared reference $\tau$ (top row) and reference F1 (bottom row) as predictors of held-out 5-shot classification performance using a leave-one-out protocol across four disease progressions. Each column represents a different held-out target task. Both metrics achieve comparable predictive accuracy (mean $\rho = 0.91$). Points represent individual foundation models.}
    \label{fig:appendix_reference_comparison}
\end{figure}

Figure~\ref{fig:appendix_reference_comparison} shows that both metrics predict held-out performance with comparable accuracy: reference $\tau$ achieves mean $\rho = 0.91$ across target tasks, while reference F1 achieves mean $\rho = 0.91$. This equivalence suggests that trajectory fidelity and classification performance both capture aspects of representation quality that generalize across tissue types. The practical value of $\tau$ lies not in superior prediction, but in its biological interpretability: trajectory fidelity directly quantifies whether models encode disease as an ordered process, providing mechanistic insight that classification accuracy alone cannot offer.

\FloatBarrier
\section{Pseudotime HistoPLUS Analysis}
\label{app:histoplus_analysis}
\subsection{Construction of Cell-Type Trajectory Plots}
\label{app:celltype_construction}

To construct Figures~\ref{fig:bio_validation}, \ref{fig:cell_composition_full}, and \ref{fig:bio_validation_other_models}, patches are sorted by diffusion pseudotime (normalized between 0 and 1), and each cell type's per-patch HistoPLUS proportion is smoothed with a centered rolling mean whose window spans 15\% of the patches in the cohort (sensitivity to this window is shown in Figure~\ref{fig:pseudotime_diagnostics}c). The smoothed proportion values are then z-scored by the mean and standard deviation of the proportion values across all patches for the cell type. The standardized curves are plotted against pseudotime.

\subsection{Sampling Diagnostics}
\label{app:pseudotime_diagnostics}

\begin{figure*}[!h]
    \centering
    \begin{minipage}[t]{0.32\textwidth}
        \centering
        \includegraphics[width=\textwidth]{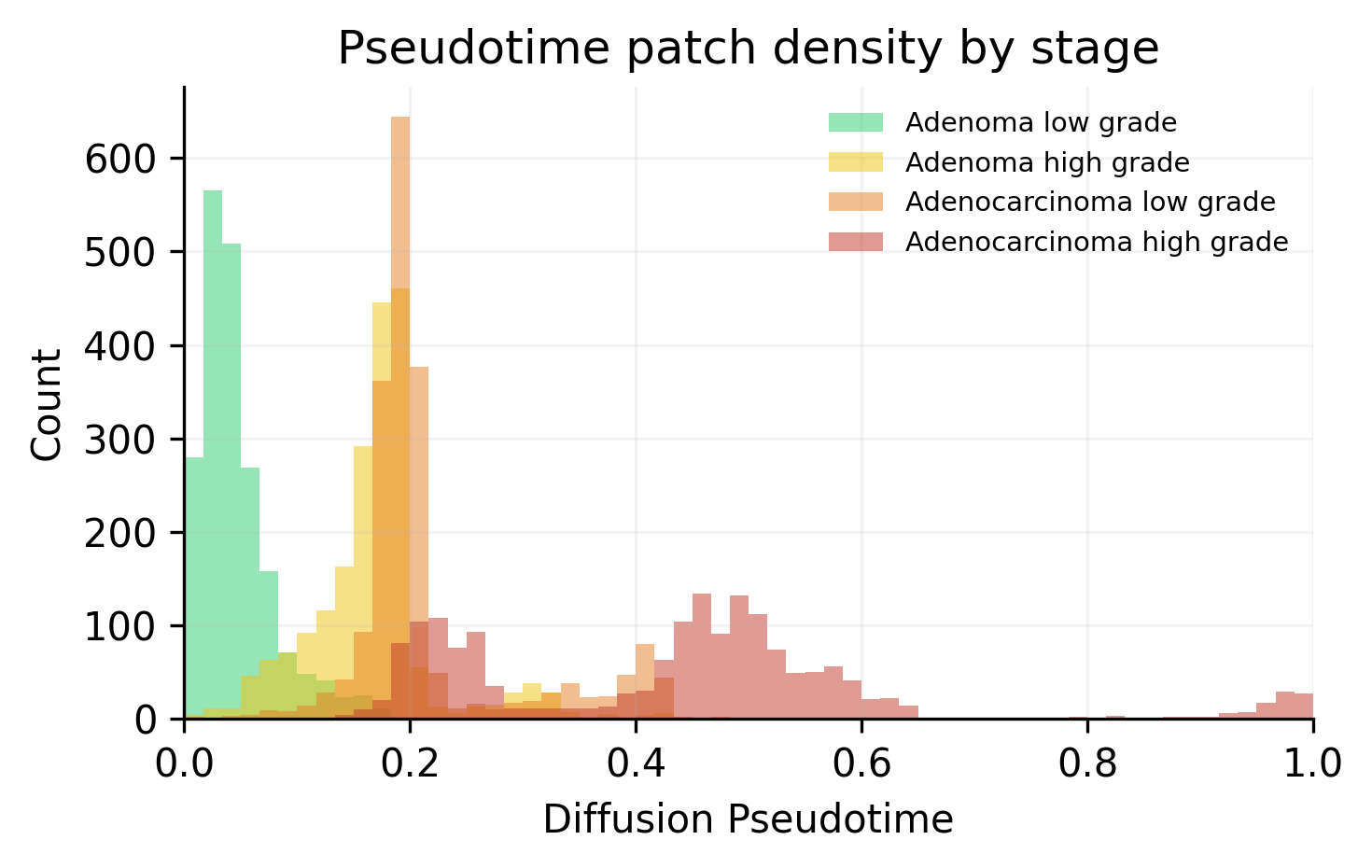}
        \caption*{\textbf{(a) Density by class.} Histogram of patches along pseudotime, colored by ground-truth class. Adjacent classes exhibit substantial overlap, indicating that pseudotime captures within-class heterogeneity rather than discrete label separation.}
    \end{minipage}
        \hfill
    \begin{minipage}[t]{0.32\textwidth}
        \centering
        \includegraphics[width=\textwidth]{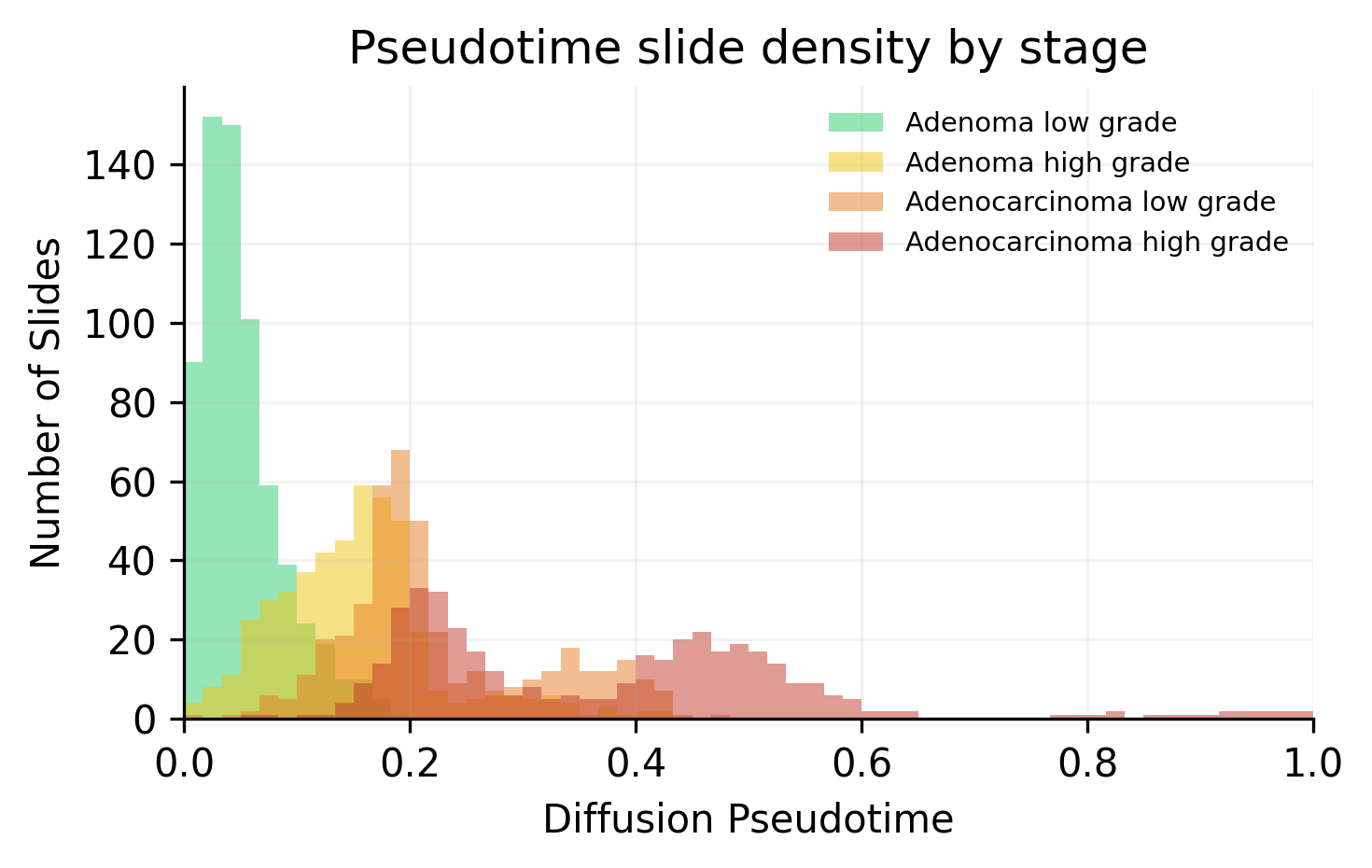}
        \caption*{\textbf{(b) Density by slide.}
        Number of distinct slides contributing at least one patch to each pseudotime bin, colored by class. Multiple slides contribute throughout the trajectory, so cell-type trends are not driven by any single slide.}
    \end{minipage}
    \hfill
    \begin{minipage}[t]{0.32\textwidth}
        \centering
        \includegraphics[width=\textwidth]{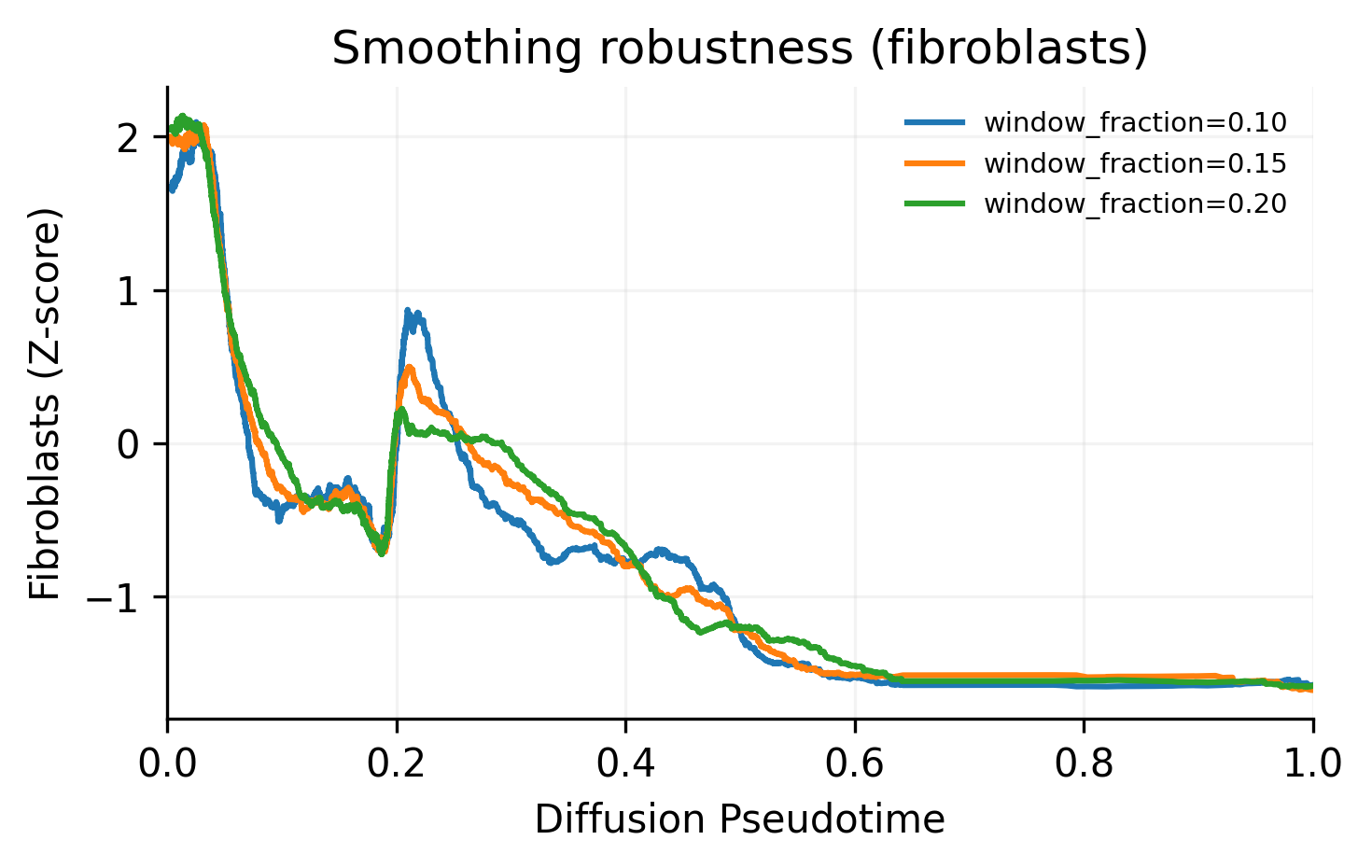}
        \caption*{\textbf{(c) Smoothing robustness.} The non-monotonic fibroblast trajectory persists across smoothing window fractions (0.10–0.20).}
    \end{minipage}
    
    \caption{\textbf{Pseudotime sampling and robustness diagnostics.}
    Patch density demonstrates strong sampling support along the trajectory, where we focus our analysis (pseudotime $< 0.6$). Local support and smoothing analyses indicate that observed cell-type trends are stable and not artifacts of smoothing choices.}
    \label{fig:pseudotime_diagnostics}
\end{figure*}

To verify that the biological trends observed along pseudotime are well supported by the underlying patch distribution and are not artifacts of smoothing, we report pseudotime density, local sampling support, and robustness of smoothed trajectories to the smoothing window fraction. This analysis is available in Figure~\ref{fig:pseudotime_diagnostics}.

\subsection{Cell Type Composition Along Pseudotime}
\label{app:cell_composition}

\begin{figure}[htbp]
    \centering
    \includegraphics[width=1.0\linewidth]{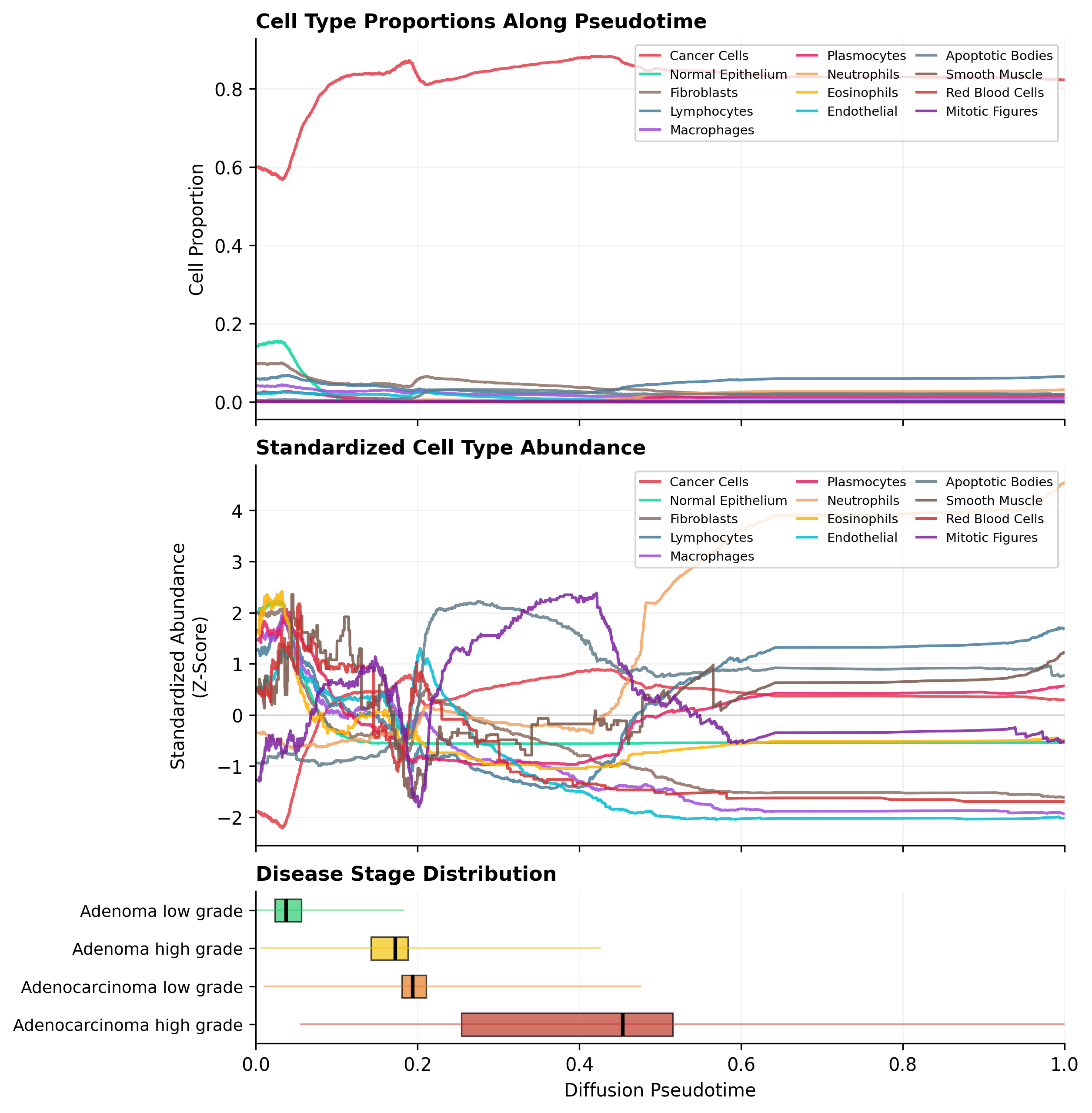}
    \caption{\textbf{Cell type composition along pseudotime.} 
    \textbf{Top:} Raw cell proportions (smoothed) for all 13 cell types across the pseudotime trajectory.
    \textbf{Middle:} Z-score standardized abundance scores. 
    \textbf{Bottom:} Disease class distribution along pseudotime, showing interquartile ranges (boxes), medians (vertical lines), and full ranges (whiskers) for each stage.}
    \label{fig:cell_composition_full}
\end{figure}

Cell type composition analysis for all 13 classes predicted by HistoPLUS is shown in Figure~\ref{fig:cell_composition_full}.

\begin{figure}[htbp]
    \centering
    \begin{minipage}[t]{0.48\textwidth}
        \centering
        \includegraphics[width=\textwidth]{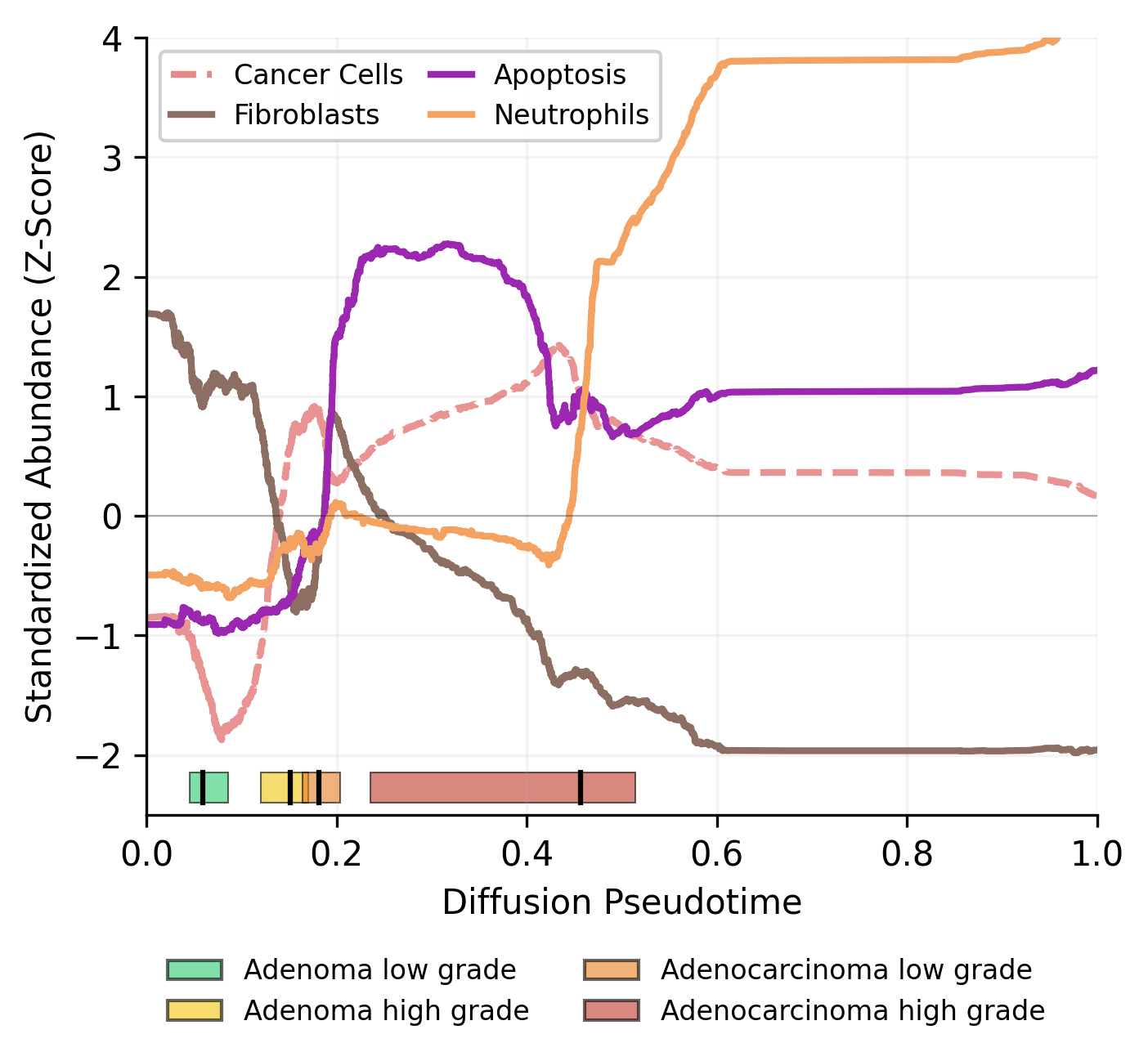}
        \caption*{\textbf{(a) Virchow-2} ($\tau = 0.73$).}
    \end{minipage}
    \hfill
    \begin{minipage}[t]{0.48\textwidth}
        \centering
        \includegraphics[width=\textwidth]{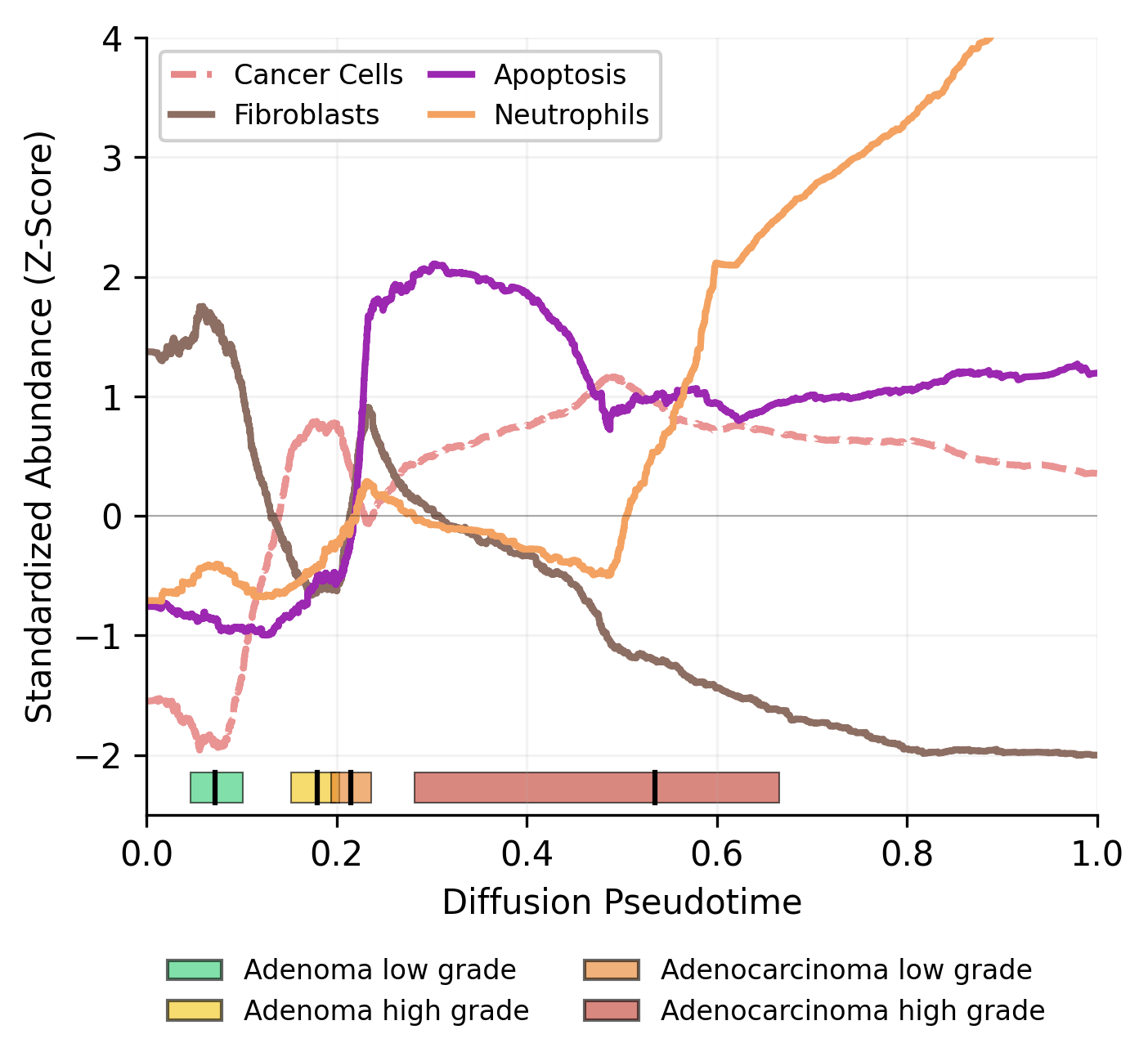}
        \caption*{\textbf{(b) Prov-GigaPath} ($\tau = 0.74$).}
    \end{minipage}
    \caption{\textbf{Biological evaluation of the learned trajectory across additional foundation models.}
    Standardized cell-type abundance along inferred pseudotime for the CRC-Conventional progression cohort, computed using (a) Virchow-2 and (b) Prov-GigaPath embeddings.}
    \label{fig:bio_validation_other_models}
\end{figure}

We additionally reproduce Figure~\ref{fig:bio_validation} for two other models,
Prov-GigaPath ($\tau = 0.74$) and Virchow-2 ($\tau = 0.73$), which achieve
trajectory fidelities comparable to UNI-2 ($\tau = 0.72$) on the CRC-Conventional
ordering. As shown in Figure~\ref{fig:bio_validation_other_models}, the
standardized cell-type abundance trends along pseudotime are qualitatively
consistent across models: fibroblast abundance exhibits the same V-shaped
pattern centered near the adenoma--adenocarcinoma transition, and neutrophil
infiltration continues to increase through the high-grade adenocarcinoma region.

\FloatBarrier
\subsection{Pseudotime: Additional Sampled Trajectory}
\label{app:traj_viz}

\begin{figure}[htbp]
    \centering
    \includegraphics[width=1.0\linewidth]{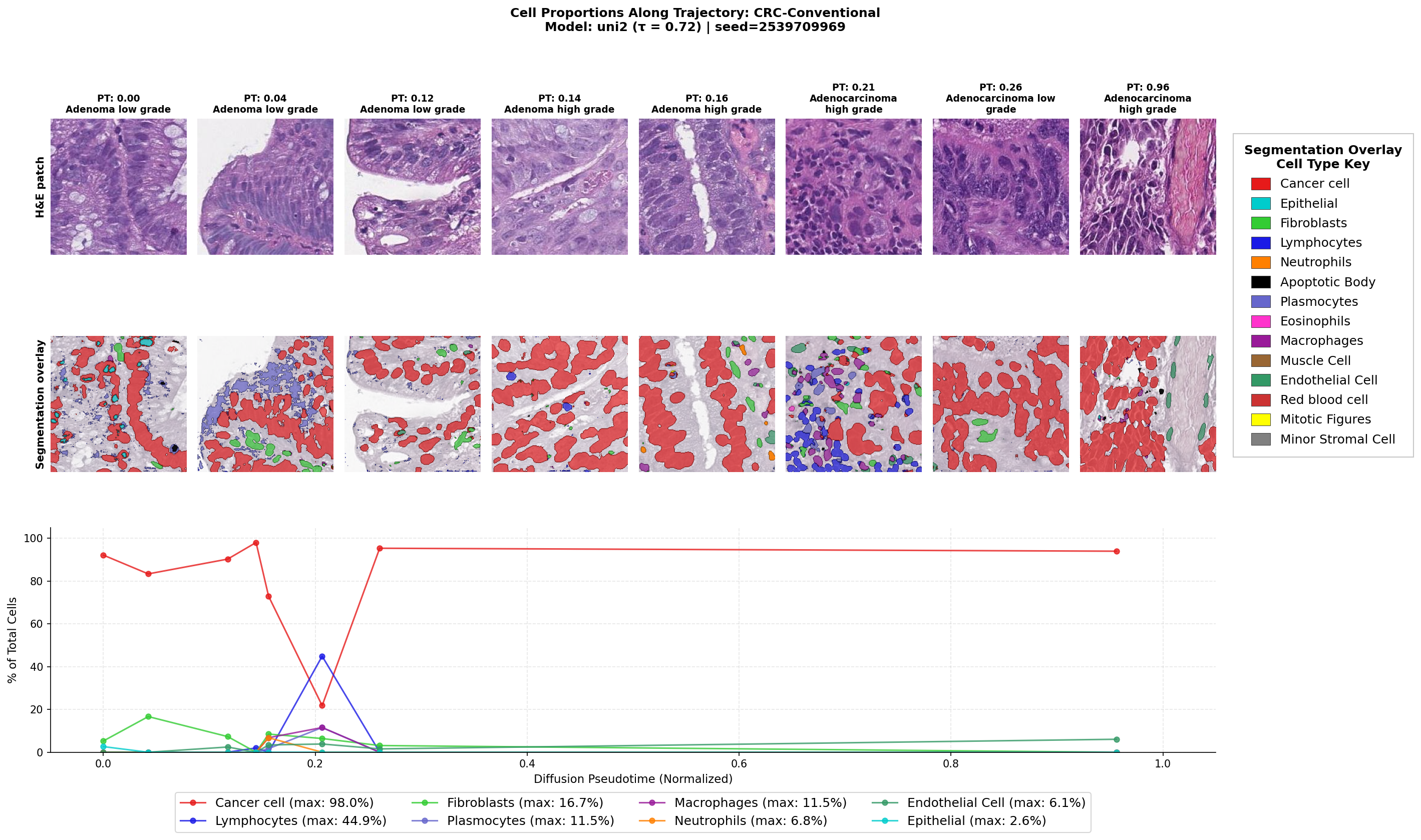}
    \caption{\textbf{Visualizing morphological disease progression via pseudotime sampling.} The top row shows representative patches sampled at increasing pseudotime points, based on UNI-2 embeddings. The second row shows the same images, with cell segmentation masks from HistoPLUS. The bottom row shows cell proportion patterns across the sampled patches. This visualization highlights the progressive morphological changes from relatively ordered tissue structures to increasingly disorganized and malignant phenotypes in CRC, captured by the model's pseudotime progression.}
    \label{fig:appendix_trajectory_sampling}
\end{figure}

Figure~\ref{fig:appendix_trajectory_sampling} shows example patches sampled across UNI-2-inferred pseudotime for CRC-Conventional. HistoPLUS segmentations are also provided for each patch.

\begin{figure}[htbp]
    \centering
    \includegraphics[width=\linewidth]{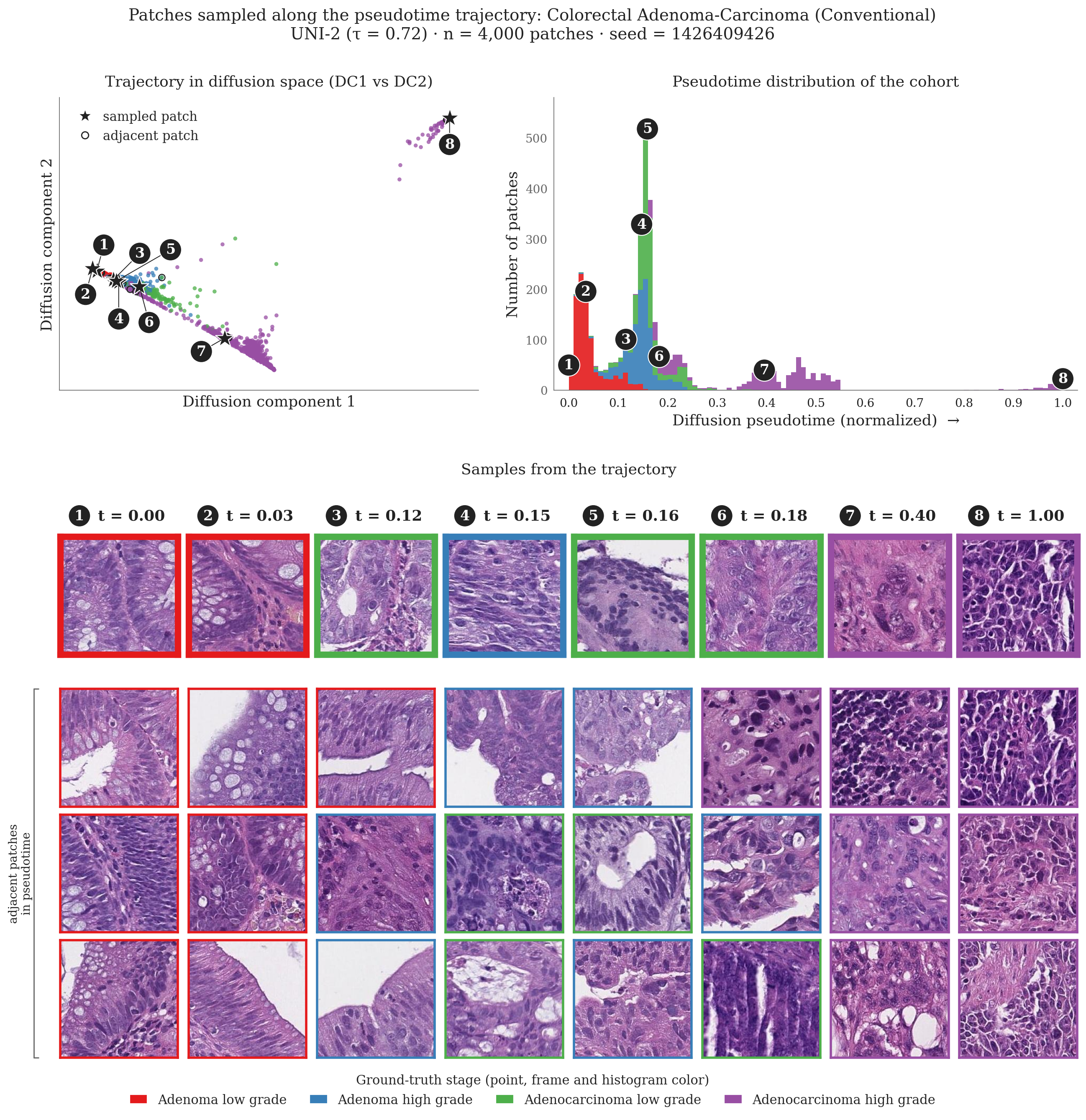}
    \caption{\textbf{Sampled patches from the CRC-Conventional trajectory and their pseudotime neighbors.} UNI-2 embeddings of the 4{,}000-patch CRC-Conventional cohort ($\tau = 0.72$). \textit{Top left:} first two diffusion components colored by ground-truth class; numbers mark eight patches sampled at evenly spaced ranks of the pseudotime ordering (first and last included). \textit{Top right:} distribution of normalized pseudotime, stacked by class, with each sampled patch marked at its pseudotime. \textit{Bottom:} the eight sampled patches (top row) and, beneath each, three patches adjacent in pseudotime drawn from distinct slides. Frame color indicates ground-truth class.}
    \label{fig:appendix_trajectory_adjacent}
\end{figure}

Figure~\ref{fig:appendix_trajectory_adjacent} provides a complementary view on the CRC-Conventional trajectory. Eight patches are drawn at evenly spaced ranks of the pseudotime ordering, and each is shown alongside three patches adjacent to it in pseudotime but drawn from distinct slides. Patches adjacent in pseudotime may fall in neighboring ground-truth classes while remaining visually similar to the sampled patch.

\section{Pseudotime Concordance with Non-Morphological Endpoints}
\label{app:tcga}

The progressions in the main text are defined by morphology. As an exploratory test of whether the inferred orderings also track progression axes not given directly by H\&E, we applied the framework to TCGA kidney renal clear cell carcinoma (KIRC) using the TCGA-UT release~\citep{komura2022universal}, which provides pathologist-annotated tumor regions tiled at $256\times256$ pixels and 0.5~$\mu$m/px. We chose KIRC because it has a relatively representative stage distribution compared to other cohorts and its availability of grade labels, which is determined by local morphology, allows a root selection consistent with the main text (medoid of the lowest-grade patches). Further, TCGA includes survival outcomes and molecular biomarkers absent from SPIDER~\citep{liu2018integrated, heath2021nci}. Restricting KIRC to primary tumors with an available grade yielded 379 patients, with up to 30 patches per patient. We measured concordance between patient-level pseudotime (mean over patches) and three endpoints: pathological stage ($n=377$), disease-specific survival (DSS; $n=371$, 81 events), and aneuploidy score, a genomic count of chromosome arms with copy-number gain or loss~\citep{taylor2018aneuploidy, cerami2012cbio, gao2013cbioportal} ($n=345$). Stage is determined by tumor size and local, nodal, and distant spread, none of which a single patch encodes directly. Pseudotime values were derived using the same DPT settings as the main text, with concordance quantified using Kendall's $\tau_b$ for stage and aneuploidy, and the C-index for DSS, to account for censoring, with significance assessed against a patient-level label-shuffled null (2{,}000 draws, one-sided). 
 
All CPath models exceed the null on every endpoint (Table~\ref{tab:tcga}), with UNI-2 and Virchow-2 strongest (e.g., DSS C-index $> 0.66$). DINOv2 is marginal, significant at one-sided $p<0.05$ on each endpoint but at the edge of the null confidence interval band. As expected, absolute concordance is lower than for the morphology-defined progressions in the main text.
 
\begin{table}[h]
\centering
\caption{\textbf{Concordance between DPT-inferred pseudotime and non-morphological endpoints on TCGA-KIRC.} Kendall's $\tau_b$ for stage and aneuploidy; C-index for DSS. Parentheses give the $z$-score against the label-shuffled null. All cells exceed the null at one-sided $p \le 0.0005$ unless marked ($^{a}$~$p=0.004$, $^{b}$~$p=0.010$, $^{c}$~$p=0.021$, $^{d}$~$p=0.031$). Null 95\% bands: stage $\pm 0.08$, aneuploidy $\pm 0.07$, DSS $[0.43, 0.57]$.}
\label{tab:tcga}
\small
\begin{tabular}{lccc}
\toprule
Model & Stage $\tau$ & Aneuploidy $\tau$ & DSS C-index \\
\midrule
UNI-2         & 0.238 (6.0) & 0.204 (5.5) & 0.671 (5.1) \\
Virchow-2     & 0.251 (6.4) & 0.180 (4.8) & 0.662 (4.7) \\
CONCH         & 0.187 (4.9) & 0.157 (4.2) & 0.626 (3.6) \\
MuSK          & 0.225 (5.7) & 0.074 (2.0)$^{c}$ & 0.627 (3.7) \\
Prov-GigaPath & 0.181 (4.7) & 0.109 (2.8)$^{a}$ & 0.591 (2.7)$^{a}$ \\
DINOv2        & 0.090 (2.4)$^{b}$ & 0.072 (1.9)$^{d}$ & 0.581 (2.4)$^{b}$ \\
\bottomrule
\end{tabular}
\end{table}

\section{Compute Resources}
\label{app:compute}

All experiments were performed on a single NVIDIA Tesla T4 GPU (16~GB VRAM) with an 8-core Intel Xeon CPU at 2.30~GHz and 29~GB of system memory. We did not train any foundation models; all analyses use publicly released pre-trained weights. Embedding extraction across the four progression cohorts and six foundation models required approximately 24 hours of GPU time. Diffusion pseudotime computation, bootstrap resampling (100 iterations), permutation analysis, and intermediate-layer emergence analyses required approximately 24 hours of additional compute. Total compute for the experiments reported is on the order of two GPU-days on a single Tesla T4.
\FloatBarrier

\end{document}